\documentclass{article}

\usepackage{microtype}
\usepackage{graphicx}
\usepackage{subfigure}
\usepackage{booktabs} 
\usepackage{hyperref}

\usepackage[accepted]{icml2024}

\usepackage{amsmath}
\usepackage{amssymb}
\usepackage{mathtools}
\usepackage{amsthm}

\usepackage{multirow}
\usepackage{pifont}       
\usepackage{bbding}       
\usepackage{fontawesome}  
\usepackage{overpic}
\usepackage{xcolor}
\usepackage{transparent}
\usepackage{enumitem}

\usepackage{dashrule}

\usepackage{silence}
\theoremstyle{plain}

\theoremstyle{definition}

\theoremstyle{remark}

\usepackage[textsize=tiny]{todonotes}

\def\eg{\emph{e.g.,~}}
\def\ie{\emph{i.e.,~}}

\newcommand{\figref}[1]{Fig.~\ref{#1}}
\newcommand{\tabref}[1]{Tab.~\ref{#1}}
\newcommand{\secref}[1]{Sec.~\ref{#1}}
\newcommand{\eqcref}[1]{Eq.~\ref{#1}}
\newcommand{\myPara}[1]{\noindent\textbf{#1}}

\newcommand{\cmark}{\ding{51}}
\newcommand{\xmark}{\ding{55}}

\definecolor{deepgreen}{rgb}{0,0.3,0.2}
\newcommand{\ourMthd}{Cascade-CLIP}

\icmltitlerunning{Cascade-CLIP: Cascaded Vision-Language Embeddings Alignment for Zero-Shot Semantic Segmentation}

\begin{document}

\twocolumn[
\icmltitle{Cascade-CLIP: Cascaded Vision-Language Embeddings Alignment for Zero-Shot Semantic Segmentation}


\begin{icmlauthorlist}
\icmlauthor{Yunheng Li}{nankai}
\icmlauthor{Zhong-Yu Li}{nankai}
\icmlauthor{Quansheng Zeng}{nankai}
\icmlauthor{Qibin Hou}{nankai,shenzhen}
\icmlauthor{Ming-Ming Cheng}{nankai,shenzhen}
\end{icmlauthorlist}

\icmlaffiliation{nankai}{VCIP, School of Computer Science, Nankai University}
\icmlaffiliation{shenzhen}{Nankai International Advanced Research Institute (Shenzhen Futian)}

\icmlcorrespondingauthor{Qibin Hou}{houqb@nankai.edu.cn}

\icmlkeywords{Zero-shot Semantic Segmentation, Multi-level visual features, Cascaded decoders}

\vskip 0.3in
]

\printAffiliationsAndNotice{}  

\graphicspath{{./images/}}




\begin{abstract}
Pre-trained vision-language models, \eg~CLIP, have been successfully applied to zero-shot semantic segmentation.
Existing CLIP-based approaches primarily utilize visual features from the last layer to align with text embeddings, while they neglect the crucial information in intermediate layers that contain rich object details. 
However, we find that directly aggregating the multi-level visual features weakens the zero-shot ability for novel classes. 
The large differences between the visual features from different layers make these features hard to align well with the text embeddings.
We resolve this problem by introducing a series of independent decoders to align the multi-level visual features with the text embeddings in a cascaded way, forming a novel but simple framework named \ourMthd{}.
Our \ourMthd{} is flexible and can be easily applied to existing zero-shot semantic segmentation methods.
Experimental results show that our simple \ourMthd~achieves superior zero-shot performance on segmentation benchmarks, like COCO-Stuff, Pascal-VOC, and Pascal-Context.
Our code is available at \url{https://github.com/HVision-NKU/Cascade-CLIP}.

\end{abstract}

\section{Introduction}
\label{sec: introduction}
Semantic segmentation, as one of the fundamental topics in computer vision, has achieved remarkable success in predicting the category of each pixel of an image~\cite{Chen_2021_CVPR,huang2021alignseg,xie2021segformer,Cheng_2022_CVPR}.
However, semantic segmentation models~\cite{Zhao_2017_CVPR,zeng2022multi} trained on closed-set annotated images are only capable of segmenting the predefined categories.
This motivates some researchers to study the zero-shot semantic segmentation models~\cite{bucher2019zero,gu2020context,han2023global}, which can segment categories that are not even present in the training images and is attracting more and more attention.

\begin{figure}[t]
    \setlength{\abovecaptionskip}{0pt}
    \centering
    \vspace{10pt}
    \begin{overpic}[width=0.47\textwidth]{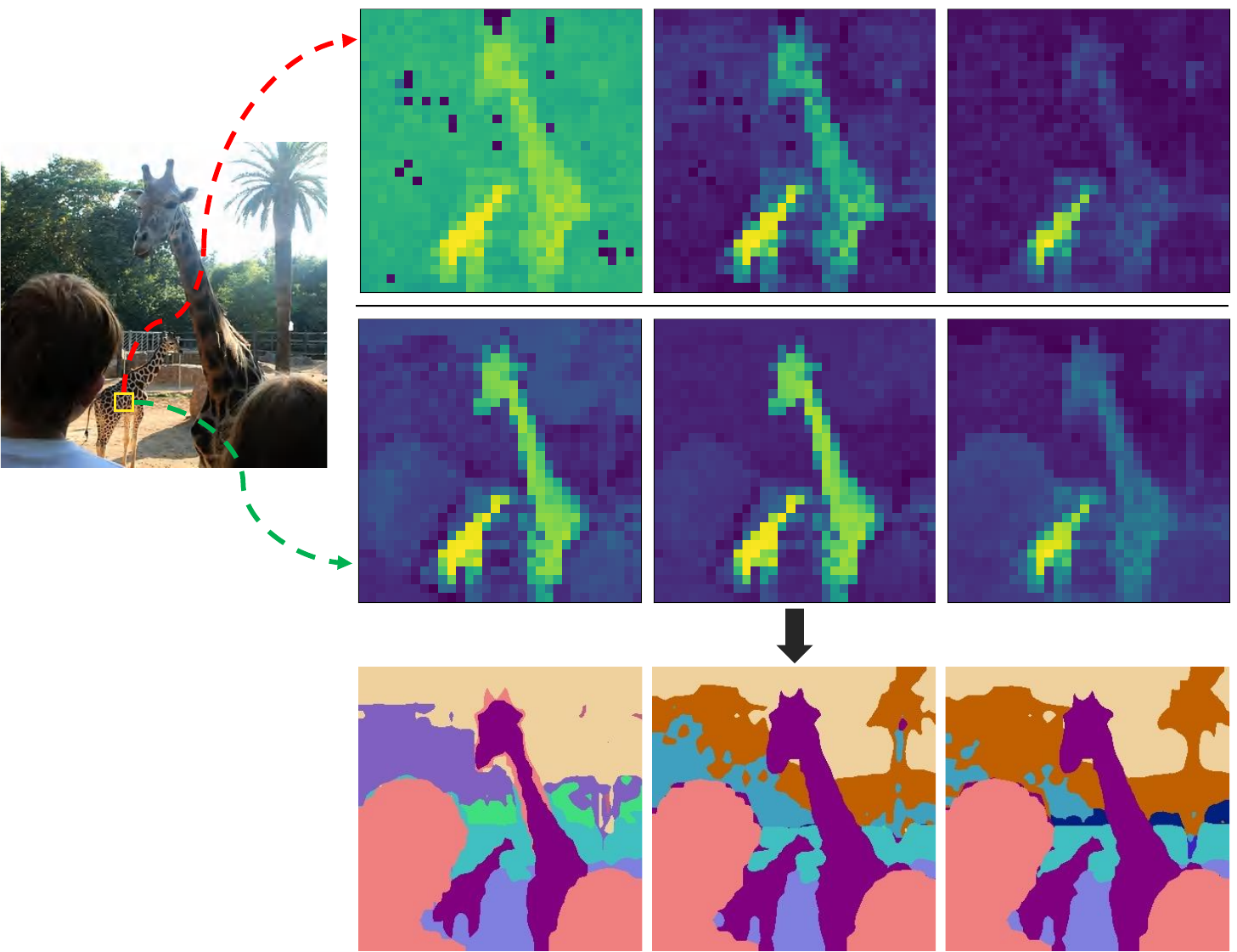}
    \put(8,65){\rotatebox{30}{\footnotesize{Original CLIP}}}
    \put(9,36){\rotatebox{-30}{\footnotesize{\ourMthd}}}
    \put(9.8,30){\rotatebox{-30}{\footnotesize{(Ours)}}}

    \put(6,11){\rotatebox{0}{\footnotesize{Cascaded}}}
    \put(2,7){\rotatebox{0}{\footnotesize{Semantic Masks}}}
    
    \put(36,77.5){\scriptsize{Layer 12}}
    \put(59,77.5){\scriptsize{Layer 11}}
    \put(83,77.5){\scriptsize{Layer 7}}

    \put(67,24){\footnotesize{Generate}}
    
    \put(33,-4){\scriptsize{Cascade\{12\}}}
    \put(53,-4){\scriptsize{Cascade\{12, 11\}}}
    \put(76,-4){\scriptsize{Cascade\{12, 11, 7\}}}
    \end{overpic}
    \vspace{2pt}
\caption{Motivation illustration of \ourMthd.
The cosine similarity map (above) indicates the visual features from the intermediate layers of CLIP~\cite{pmlr-v139-radford21a} layers can capture richer local object details compared to the last one (Layer 12).
}
\label{Image_cosine_similarity3}
\vskip -0.2in
\end{figure}

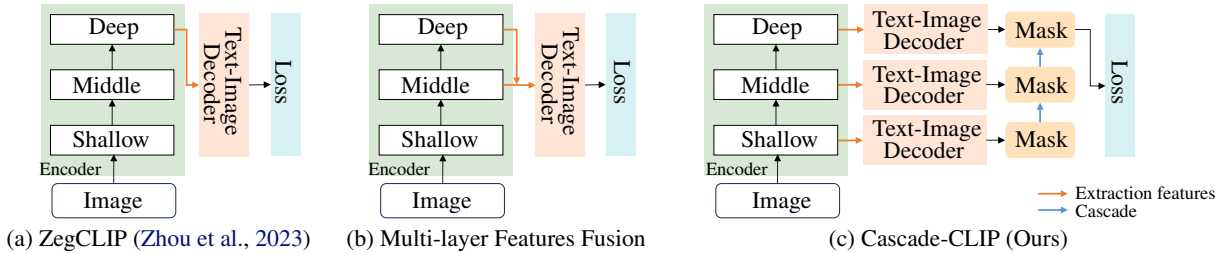
\begin{figure*}[!t]
\centering
    \setlength{\abovecaptionskip}{0pt}
    \vspace{4pt}
    \begin{overpic}[width=0.85\textwidth]{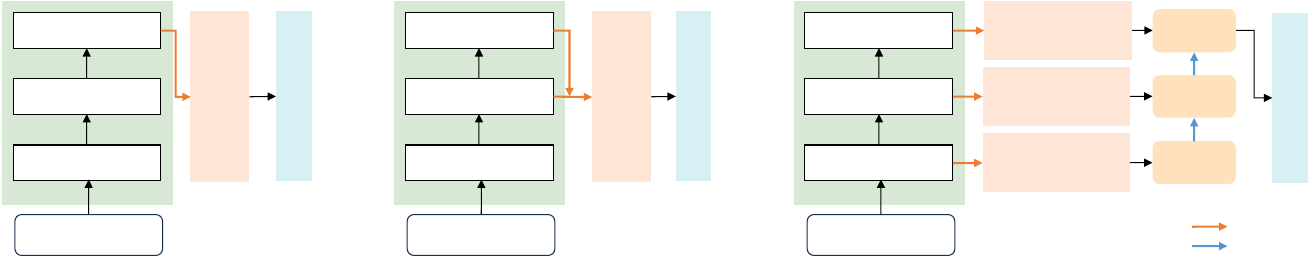}
    \put(-3,-2.5){\footnotesize{(a) ZegCLIP~\cite{Zhou_2023_CVPR}}}
    \put(4,1){\small{Image}}
    \put(0,4){\scriptsize{Encoder}}
    \put(3,6.5){\small{Shallow}}
    \put(3,11.5){\small{Middle}}
    \put(4.5,16.8){\small{Deep}}
    \put(15,16){\rotatebox{-90}{\small{Decoder}}}
    \put(16.5,17.5){\rotatebox{-90}{\small{Text-Image}}}
    \put(21.5,14.5){\rotatebox{-90}{\small{Loss}}}

    \put(28,-2.5){\footnotesize{(b) Multi-layer Features Fusion}}
    \put(34,1){\small{Image}}
    \put(30,4){\scriptsize{Encoder}}
    \put(33,6.5){\small{Shallow}}
    \put(33,11.5){\small{Middle}}
    \put(34.5,16.8){\small{Deep}}
    \put(45.8,16){\rotatebox{-90}{\small{Decoder}}}
    \put(47.3,17,5){\rotatebox{-90}{\small{Text-Image}}}
    \put(52.4,14.5){\rotatebox{-90}{\small{Loss}}}

    \put(72,-2.5){\footnotesize{(c) \ourMthd~(Ours)}}
    \put(65,1){\small{Image}}
    \put(60.8,4){\scriptsize{Encoder}}
    \put(64,6.5){\small{Shallow}}
    \put(64,11.5){\small{Middle}}
    \put(65,16.8){\small{Deep}}
    
    \put(76,17.5){{\small{Text-Image}}}
    \put(77,15.5){{\small{Decoder}}}
    \put(76,12.5){{\small{Text-Image}}}
    \put(77,10.5){{\small{Decoder}}}
    \put(76,7.5){{\small{Text-Image}}}
    \put(77,5.5){{\small{Decoder}}}
    \put(97.8,14.5){\rotatebox{-90}{\small{Loss}}}
    
    \put(89,6.5){{\small{Mask}}}
    \put(89,11.5){{\small{Mask}}}
    \put(89,16.5){{\small{Mask}}}


    \put(94.5,1.8){{\scriptsize{Extraction features}}}
    \put(94.5,0.1){{\scriptsize{Cascade}}}

    \end{overpic}
    \vspace{7pt}
\caption{Three zero-shot segmentation approaches based on CLIP.
(a) ZegCLIP relies on the \textbf{last-layer} visual features without considering information from intermediate layers.
(b) Inspired by SegFormer~\cite{xie2021segformer}, we fuse both \textbf{intermediate-} and \textbf{last-layers} features to enhance feature representation, yet this integration disrupts the correlation between text and visual features.
%
(c) To alleviate this issue, our \ourMthd~\textbf{separats} the image encoder and \textbf{aligns} independent text-image decoders for deep features and middle features respectively, and finally \textbf{cascades} the segmentation results.
}
\label{Introduction}
\vskip -0.2in
\end{figure*}

Recently, benefiting from the impressive zero-shot capability at the image level, large-scale visual-language pre-training models, typified by CLIP~\cite{pmlr-v139-radford21a}, have been considered in zero-shot semantic segmentation.
Nevertheless, directly applying CLIP to the zero-shot semantic segmentation task is ineffective as it requires dense pixel/region-wise predictions.
Two-stage methods~\cite{xu2021simple,ding2022decoupling} solve the aforementioned issue via generating region proposals by trained proposal generator and feeding the cropped masked regions to CLIP for zero-shot classification.
Although this paradigm retains CLIP's image-level zero-shot ability well, it introduces high computational cost.
The one-stage line~\cite{Zhou_2023_CVPR,xu2023spectral} produces pixel-wise segmentation by matching text embeddings and pixel-level features extracted from the last layer of CLIP's visual encoder, achieving a good balance between efficiency and effectiveness.
However, these methods exhibit weaknesses in segmenting object details, especially the boundaries of the semantic objects.

Building on insights from the closed-set segmentation methods~\cite{zheng2021rethinking,hou2020strip,xie2021segformer,guo2022segnext}, a viable solution for capturing rich local details is to aggregate multi-level features from the encoder to improve the coarse segmentation results.
For the CLIP model, we observe that the visual features extracted from the intermediate layers contain rich object details as illustrated in~\figref{Image_cosine_similarity3}.
Nevertheless, directly fusing the multi-level features produces unsatisfactory results.
As shown in \tabref{tab: Tab-Single-Multi-Ours}, a straightforward fusion of the middle-layers and the last-layer features~(\figref{Introduction}(b)) degrades the performance compared to the baseline model~(\figref{Introduction}(a)).
The baseline's success lies in effectively leveraging the pre-trained correlations in CLIP between the last-layer visual features and the text embeddings.
However, the fusion of multi-level features disrupts these original visual-language correlations due to the significant disparity between the middle-layer and last-layer features, weakening CLIP's zero-shot capability on unseen classes.
Moreover, after feature fusion, the differences between features also disrupt the pre-trained visual representations, further increasing the difficulty of aligning visual features with the text embeddings during fine-tuning.

\begin{table}[t]
    \centering
    \vskip -0.05in
    \small
    \setlength{\abovecaptionskip}{5pt}
    \setlength{\tabcolsep}{0.4mm}
    \caption{Comparisons with different methods on COCO-Stuff 164K, and PASCAL VOC 2012 datasets\footnotemark[1].
    %
    %
    Straightforward fusion of multi-layer features leads to performance degradation.
    %
    }
    \label{tab: Tab-Single-Multi-Ours}
    \begin{tabular}{ccccccccc} \toprule
    & \multicolumn{3}{c}{COCO-Stuff 164K}&\multicolumn{3}{c}{PASCAL VOC 2012} \\  
    \cmidrule(lr){2-4}  \cmidrule(lr){5-7}
    Methods & mIoU$^S$& mIoU$^U$& hIoU& mIoU$^S$& mIoU$^U$& hIoU \\ \midrule
    ZegCLIP (Baseline) & 40.1 & 39.5 & 39.8 & 90.5& 78.3& 84.0 \\
    Multi-layer Fusion& 37.7& 39.0& 38.4& 91.2& 75.0 &82.3\\
    \ourMthd &\textbf{41.1} &\textbf{43.4}& \textbf{42.2} & \textbf{92.7} & \textbf{83.1} & \textbf{87.7} \\ 
    \bottomrule
    \end{tabular}
    \vskip -0.2in
\end{table}

\footnotetext[1]{mIoU$^S$ and mIoU$^U$ represent the mean Intersection over Union (\%) of seen and unseen classes, respectively. hIoU denotes the harmonic mean IoU score among seen and unseen classes.}

In this paper, we renovate the way of aligning visual and text embeddings and propose \ourMthd, a multi-level framework that can better leverage the diverse visual features from CLIP and enhance the transferability to novel classes.
Specifically, \ourMthd{} splits the visual encoder into multiple stages, ensuring a little variation in the features within each stage.
Each stage is equipped with an independent text-image decoder, employing distinct text embeddings to align the multi-level visual features better and build better vision-language correlation.
In this way, we can integrate complementary multi-level semantic masks from the visual encoder to enhance the segmentation results as demonstrated in \figref{Image_cosine_similarity3} (bottom row).

By exploiting multi-level features, for the first time, we demonstrate that our \ourMthd{} can largely improve the image-to-pixel adaptability of CLIP in zero-shot semantic segmentation.
In addition, \ourMthd{} is also flexible and can be seamlessly utilized in existing state-of-the-art methods, such as ZegCLIP~\cite{Zhou_2023_CVPR} and SPT-SEG~\cite{xu2023spectral}, to lift their performance on three commonly used zero-shot segmentation benchmarks. 
In particular, thanks to the cascaded vision-language alignment, our method performs especially well for unseen classes, reflecting strong adaptability.
The contributions can be summarized as follows:
\begin{itemize}[topsep=0pt, partopsep=0pt, itemsep=1pt, parsep=0.5pt]
    \item We unveil that visual features from the intermediate layers of CLIP contain rich local information about objects. However,
    simply fusing the multi-level visual features weakens CLIP's zero-shot capability. 
    \item We propose \ourMthd{}, a flexible cascaded vision-language embedding alignment framework that can effectively leverage the multi-level visual features from CLIP to improve the transferability for novel classes.
    \item Extensive experiments demonstrate the effectiveness of our \ourMthd~in zero-shot semantic segmentation on three widely-used benchmarks.
\end{itemize}

\section{Related work}
\subsection{Pre-trained Vision Language Models}
Large-scale vision-language models~\cite{pmlr-v139-jia21b,pmlr-v139-kim21k,pmlr-v139-radford21a} pre-trained with web-scale image-text pairs have made great progress in aligning image and text embeddings and achieved strong zero/few-shot generalization capabilities. 
For example, one of the most popular vision-language models, CLIP~\cite{pmlr-v139-radford21a}, is trained contrastively using 400 million image-text pairs. 
Due to its zero-shot recognition capabilities and its simplicity, CLIP has been widely adapted to various down-stream tasks, 
such as zero-shot visual recognition~\cite{Khattak_2023_ICCV}, dense prediction~\cite{Rao_2022_CVPR}, object detection~\cite{gu2021open}, and visual referring expression~\cite{Wang_2022_CVPR}. 
This work explores how to efficiently transfer CLIP's powerful generalization ability from images to pixel-level classification.

\subsection{Zero-shot Semantic Segmentation}
Zero-shot semantic segmentation performs a pixel-level classification that includes unseen categories during training. 
Previous works, such as SPNet~\cite{xian2019semantic}, ZS3~\cite{bucher2019zero}, CaGNet~\cite{gu2020context}, SIGN~\cite{cheng2021sign}, JoEm~\cite{baek2021exploiting}, and STRICT~\cite{Pastore_2021_CVPR}, focus on learning a mapping between visual and semantic spaces to improve the generalization ability of the semantic mapping from the seen classes to the unseen ones.
Recent approaches mostly employ large-scale visual-language models (\eg CLIP~\cite{pmlr-v139-radford21a} and ALIGN~\cite{pmlr-v139-jia21b}) with powerful zero-shot classification ability for zero-shot semantic segmentation.
Some training-free methods, such as ReCo~\cite{shin2022reco} and CaR~\cite{sun2023clip}, directly use CLIP to perform zero-shot semantic segmentation. 
Other methods, like MaskCLIP+~\cite{zhou2022extract}, apply CLIP to generate pseudo-annotations for unseen classes to train existing segmentation models, but the requirement for unseen class names constrains it.
%
%
To relieve this limitation, some works, like ZegFormer~\cite{ding2022decoupling}, Zsseg~\cite{xu2021simple}, FreeSeg~\cite{Qin_2023_CVPR}, and DeOP~\cite{han2023open}, decouple the zero-shot semantic segmentation into the class-agnostic mask generation process and the mask category classification process using CLIP. 
While they retain the zero-shot capability of CLIP at the image level, the computational cost inevitably increases due to the introduction of proposal generators.

Instead of using heavy proposal generators, ZegCLIP~\cite{Zhou_2023_CVPR} introduces a lightweight decoder to match the text embeddings against the visual embeddings extracted from CLIP.
Similarly, SPT-SEG~\cite{xu2023spectral} enhances the CLIP's semantic understanding ability by integrating spectral information.
Although the above methods have successfully translated CLIP's image classification into pixel segmentation, there is still a large room for improvement.
Unlike previous works, we take a new look at zero-shot semantic segmentation by investigating the role of the features from the intermediate layers in the visual encoder.

\begin{figure*}[!t]
    \setlength{\abovecaptionskip}{0pt}
    \centering
        \begin{minipage}[c]{0.8\linewidth}
        \begin{overpic}[width=1\textwidth]{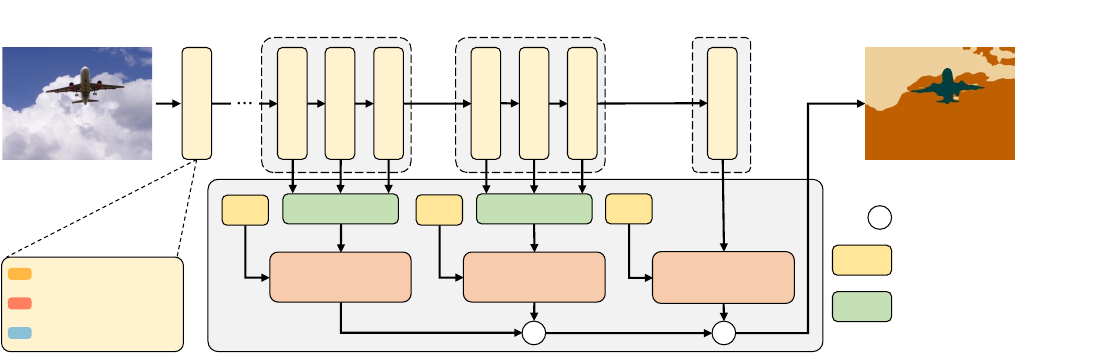}

        \put(16,29.5){\small{CLIP Frozen Visual Encoder with Learnable Tokens}}
        \put(78,29.5){\small{Segmentation Map}}

        \put(17,25.5){\rotatebox{-90}{{Block}}}

        \put(3.3,6.8){\scriptsize{[CLS] token}}
        \put(3.3,3.9){\scriptsize{Learnable tokens}}
        \put(3.3,1.2){\scriptsize{Patch tokens}}
        
        \put(25.8,25.5){\rotatebox{-90}{{Block}}}
        \put(30.2,25.5){\rotatebox{-90}{{Block}}}
        \put(34.6,25.5){\rotatebox{-90}{{Block}}}

        \put(43.5,25.5){\rotatebox{-90}{{Block}}}
        \put(48,25.5){\rotatebox{-90}{{Block}}}
        \put(52.4,25.5){\rotatebox{-90}{{Block}}}
        
        \put(65.2,25.5){\rotatebox{-90}{{Block}}}

        \put(21.2,12.2){\small{$\mathbf{\hat T}_1$}}
        \put(39,12.2){\small{$\mathbf{\hat T}_2$}}
        \put(56.3,12.2){\small{$\mathbf{\hat T}_3$}}
        
        \put(29,12.3){\small{NGA}}
        \put(28,10){\scriptsize{$\mathbf{Z_1}$}}
        \put(47,12.3){\small{NGA}}
        \put(46,10){\scriptsize{$\mathbf{Z_2}$}}
        \put(63,10){\scriptsize{$\mathbf{Z_3}$}}

        \put(19.5,2.5){\small{\textbf{Cascaded}}}
        \put(19.5,0.5){\small{\textbf{Decoders}}}

        \put(26,7){\small{Text-Image}}
        \put(27.5,5){\small{Decoder}}
        \put(43.8,7){\small{Text-Image}}
        \put(44.3,5){\small{Decoder}}
        \put(61,7){\small{Text-Image}}
        \put(62.5,5){\small{Decoder}}


        \put(48,1.3){\scriptsize{$+$}}
        \put(65.4,1.3){\scriptsize{$+$}}
        
        \put(79.7,11.8){\scriptsize{$+$}}
        \put(82,11.5){\footnotesize{Cascade mask}}
        \put(78,7.3){\small{$\mathbf{\hat T}$}}
        \put(82,7.3){\footnotesize{Text embedding}}
        \put(76.5,3.3){\small{NGA}}
        \put(82,3){\footnotesize{Neighborhood}}
        \put(79,.5){\footnotesize{Gaussian aggregation}}
        
        \put(99.5,0){\tikz\draw[dashed,line width = 0.5pt](0,0)--(0,4.3);}

        \end{overpic}

        \end{minipage}
        \begin{minipage}[c]{0.14\linewidth}
        \flushright
            \begin{overpic}[width=1\textwidth]{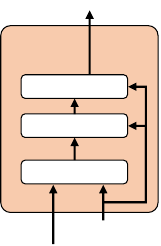}
            \put(2,97.5){\scriptsize{Mask ($\mathcal{M}\in\mathbb{R}^{C \times N}$)}}

            \put(2,82){\scriptsize{\textbf{Text-Image}}}
            \put(6,76){\scriptsize{\textbf{Decoder}}}

            \put(10,27.5){\scriptsize{Multihead Attn}}
            \put(10,47){\scriptsize{Multihead Attn}}
            \put(10,63.2){\scriptsize{Multihead Attn}}

            \put(18.5,-7){\scriptsize{$\mathbf{\hat T}\in\mathbb{R}^{C \times 2d}$}}
            \put(40,5){\scriptsize{$\mathbf{Z}\in\mathbb{R}^{N \times d}$}}

            \end{overpic}
        \end{minipage}
    \caption{Architecture of our \ourMthd.
    The CLIP visual encoder is divided into multiple stages.
    Then, we employ the NGA module to aggregate features of blocks within each stage and assign an independent text-image decoder for aggregated visual features and non-sharing text embeddings.
    In the text-image decoder (right part of the figure), the segmentation mask could be calculated by the scaled dot product attention via the Multihead Attention (Attn) layers, inspired by~\cite{zhang2022segvit}.
    Finally, we combine the multi-level semantic masks produced by different cascaded decoders to enhance segmentation predictions.
    %
    %
    (Please refer to~\secref{sec: NGA} for details.)
    }
    \label{Framework}
    \vskip -0.1in
    \end{figure*}
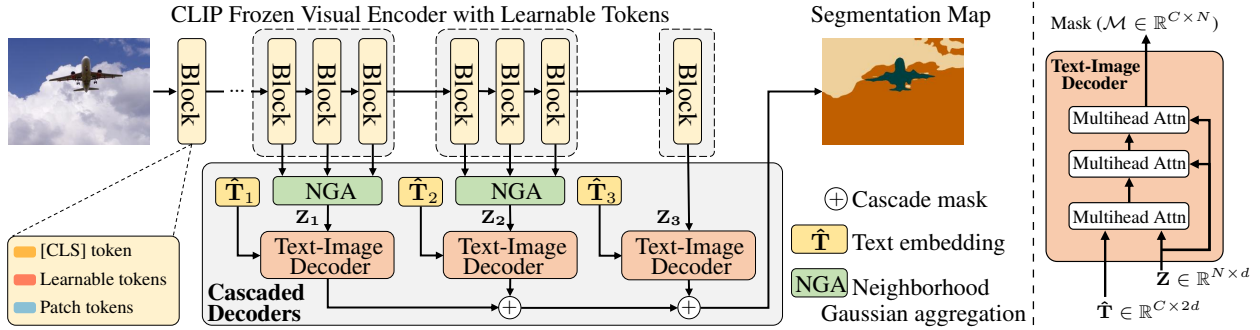  

\section{Method}

The zero-shot semantic segmentation task~\cite{bucher2019zero,Zhou_2023_CVPR} aims to segment both the seen classes $\mathcal{C}$ and unseen classes $\hat{\mathcal{C}}$ after training on a dataset with the seen part of available pixel annotations.
Normally, $\mathcal{C} \cap \hat{\mathcal{C}}=\varnothing$ and the labels of $\hat{\mathcal{C}}$ are unavailable while training.
The key problem is to retain the ability to identify unseen classes when training on the seen classes.

\subsection{Revisiting ZegCLIP} \label{sec:zegclip}
Recent zero-shot semantic segmentation approaches~\cite{Zhou_2023_CVPR,xu2023spectral} are mostly based on the one-stage scheme due to its high efficiency and good performance.
Here, we revisit the ZegCLIP work~\cite{Zhou_2023_CVPR} as our baseline.

As illustrated in~\figref{Introduction}(a), ZegCLIP~\cite{Zhou_2023_CVPR} first extracts CLIP's text embeddings of $C$ classes as $\mathbf{T}=\left[\mathbf{t}_1, \mathbf{t}_2, \ldots, \mathbf{t}_C\right] \in \mathbb{R}^{C \times d}$
and CLIP's visual features of an image as [CLS] token  $\mathbf{g}\in \mathbb{R}^{1 \times d}$ and patch tokens $\mathbf{H}\in \mathbb{R}^{N \times d}$, where $d$ is the feature dimension of the CLIP model, and $N$ is the number of patch tokens.
$C$ is the number of classes with $C = \left|\mathcal{C}\right|$ during training and $C =|\mathcal{C} \cup \hat{\mathcal{C}}|$ during inference.
To avoid overfitting, a relationship descriptor, denoted as
$\mathbf{\hat T} = \operatorname{concat}(\mathbf{T} \odot \mathbf{g}, \mathbf{T}) \in \mathbb{R}^{C \times 2d}$, where $\odot$ and $\operatorname{concat}$ are the Hadamard product and concatenation, 
is employed by~\cite{Zhou_2023_CVPR} instead of $\mathbf{T}$.
Then, the semantic masks $\mathbf{M} \in \mathbb{R}^{C \times N}$ can be generated in the text-image decoder by measuring the similarity between the text embeddings $\mathbf{\hat T}$ and the visual features $\mathbf{H}$. The whole process can be represented as follows:
\begin{equation}
    \begin{split}
    \mathbf{M}&= \operatorname{Softmax}(\mathcal{D}(\phi_q(\mathbf{\hat T}), \phi_k(\mathbf{H}))),\\
    \end{split}
\label{eq3.1-1}
\end{equation} 
where $\mathcal{D}(\cdot)$ denotes the text-image decoder, as illustrated in the right part of~\figref{Framework}.
$\phi_q$ and $\phi_k$ are two linear projections that align the feature dimension of $\mathbf{\hat T}$ and $\mathbf{H}$.

Since the visual features are only extracted from the last layer of the visual encoder, previous methods often cannot identify the boundaries of the semantic objects well. 
This is because the deep features carry high-level semantic global features as shown in~\figref{Image_cosine_similarity3} but less low-level local details 
compared to the intermediate layers, which we will pay close attention to in this paper. 

\begin{figure}[!t]
    \setlength{\abovecaptionskip}{0pt}
\centering
    \begin{overpic}[width=0.47\textwidth]{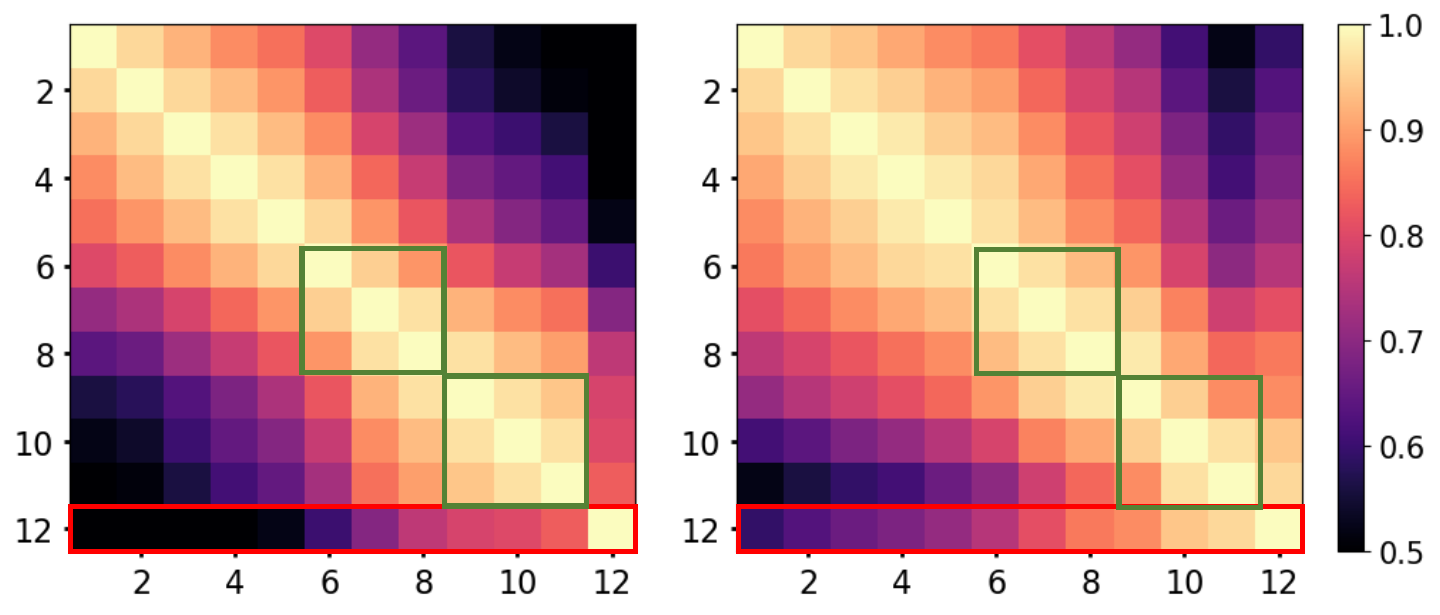}
    \put(10,-3){\footnotesize{(a) Original CLIP}}
    \put(52,-3){\footnotesize{(b) \ourMthd~(Ours)}}
    \end{overpic}
\caption{Centered kernel alignment heatmap~\cite{kornblith2019similarity} between layers of (a) Original CLIP and (b) \ourMthd~(Ours).
The last row (\textcolor{red}{red} box) shows the similarity between features from the last layer and other layers.
The \textcolor{deepgreen}{green} box illustrates the similarity between adjacent layers.
}
\label{image_encoder_cka}
\vskip -0.2in
\end{figure}

\subsection{Motivation}
\label{sec: delving}
Multi-level features are commonly used in closed-set segmentation models~\cite{zheng2021rethinking,xie2021segformer} to sharpen the object segmentation details.
Our analysis in~\secref{sec: introduction} also reveals that features from the middle layers of CLIP~\cite{pmlr-v139-radford21a} can capture rich local object details.
This motivates us to investigate how to effectively take advantage of these distinct features to enhance CLIP's transferability to novel classes, which has been omitted by previous works.
However, simply aggregating the multi-level visual features as done in~\figref{Introduction}(b) degrades the segmentation performance.
To analyze the reasons for the performance degradation, we attempt to visualize the centered kernel alignment map~\cite{kornblith2019similarity} of the visual features of CLIP as shown in~\figref{image_encoder_cka}(a), which measures the similarity between different layers.
We observe a substantial dissimilarity between the shallow and deep features, and the difference increases with network depth.
This indicates that directly integrating the multi-level intermediate features into the last ones may break the alignment of vision-language embedding in pre-trained CLIP due to the substantial differences, thereby weakening the zero-shot capability of CLIP.

Given the above analysis, we intend to investigate how to effectively leverage the intermediate features with rich local details to improve zero-shot segmentation.
To address this challenge, we propose two strategies, namely Cascaded Vision-Language Embedding Alignment and Neighborhood Gaussian Aggregation to better align the multi-level visual features with the text embeddings.
These strategies aim to reduce the feature differences between different layers so that the mid-level visual features can be well aligned with the text embeddings and complement deep-level features, improving the ability of zero-shot segmentation.

\subsection{Cascaded Alignment Framework}

\vspace{-2pt}
The overview of the proposed \ourMthd~framework is illustrated in~\figref{Framework}.
Basically, the visual encoder of CLIP is split into multiple stages to extract multi-level visual features, with a little variation for the features in each stage.
Then, to better build vision-language correlations during fine-tuning, we assign each stage of the visual encoder an independent text-image decoder considering the feature differences between various stages. 
The decoder is similar to the one mentioned in~\secref{sec:zegclip}. 
Finally, the refined results are generated by cascading complementary segmentation masks from various stages.

To be specific, let $\mathbf{H}_l$ denote the patch tokens of the $l$-th Transformer block.
For ViT-B, the number of blocks should be 12.
First, we separate the CLIP's visual encoder into $S$ stages, each containing a group of Transformer blocks.
In each stage, \eg $s$-th stage, to better leverage the multi-level features from different Transformer blocks, we introduce a Neighborhood Gaussian Aggregation (NGA) module to aggregate these features, resulting in aggregated features $\mathbf{Z}_s$.
We will describe the NGA module in detail later.
Then, for the output from the $s$-th stage $\mathbf{Z}_s$, we associate a corresponding text embedding $\mathbf{\hat T}_s$, acquired through linear projection from $\mathbf{\hat T}$.
Subsequently, $\mathbf{Z}_s$ and $\mathbf{\hat T}_s$ are fed into an independent text-image decoder to generate semantic masks.
Finally, we replace the single semantic masks in~\eqcref{eq3.1-1} by combining all semantic masks generated by multiple stages as follows:
\begin{equation}
    \begin{split}
        \mathbf{M}_S=\operatorname{Softmax}\left(\sum\nolimits_{s=1}^S \mathcal{D}_s(\mathbf{\hat T}_s, \mathbf{Z}_s)\right),
    \end{split}
    \label{eq3}
\end{equation}
where $\mathcal{D}_s(\cdot)$ denotes the $s$-th text-image decoder.
Here, we use an element-wise summation operation, which can be regarded as an ensemble of the outputs from multiple cascaded decoders.

As shown in~\figref{Framework}, the vision-language alignment process can be applied multiple times to different blocks in a cascaded way.
In practice, we do not append the text-image encoder to the shallow Transformer blocks because the shallow features contain little semantic information.
Our experiment in~\secref{exp: ablation} will show how to split the visual encoders to take advantage of the multi-level visual features.

\newcommand{\loss}[1]{\mathcal{L}^{\text{#1}}}
\newcommand{\diff}{\mathbf{Y}, \mathcal{M}}

\myPara{Loss function with cascaded masks.}
Given the text-image decoder $\mathcal{D}_s(\cdot)$ from the $s$-th stage, let $\mathcal{M}_s=\mathcal{D}_s(\mathbf{\hat T}_s, \mathbf{Z}_s)$ be the predicted segmentation mask.
$\mathcal{M} = \sum_{s=1}^{S}\mathcal{M}_s$ is the multi-level cascaded mask.
The objective loss function $\mathcal{L}^{\text{pixel}}$ is defined as:
\begin{equation}
  \loss{pixel} = \alpha \loss{dice}(\diff) + \beta \loss{focal}(\diff),
  \label{eq3.2-2}
\end{equation} 
where $\mathcal{L}^{\text {dice}}$ and $\mathcal{L}^{\text {focal}}$ are the dice loss~\cite{milletari2016v} and focal loss~\cite{lin2017focal} with $\operatorname{Sigmoid}$ as activation function, respectively.
$\mathbf{Y}$ is the ground truth.
$\{\alpha,\beta\}$ are two weights with the default values of $\{1,100\}$, respectively.

To better align the intermediate visual features with the text embeddings, we employ visual prompt tuning~\cite{zhou2022learning,ding2022decoupling} by introducing learnable tokens onto the visual features of each block in the frozen encoder.
During visual prompt tuning, the cascaded alignment manner enables the gradients to be directly back-propagated to the middle layers of the visual encoder.
This can promote the alignment of mid-layer features and text embeddings, substantially enhancing the similarity across different layers.
We illustrate this in~\figref{image_encoder_cka}(b), which reflects a clear difference from~\figref{image_encoder_cka}(a).

\myPara{Neighborhood Gaussian aggregation.}
\label{sec: NGA}
To better utilize the potential of the features from each Transformer block, 
we propose the Neighborhood Gaussian Aggregation (NGA) module to fuse the multi-level features within each stage.
Based on the analysis in~\secref{sec: delving} and the illustration in~\figref{image_encoder_cka}(b), we observe a gradual decline in feature similarity across layers with increasing distance.
Thus, we propose to assign blocks with distinct Gaussian weights during feature fusion based on their relative neighborhood distances. 
Furthermore, these weights are trainable concerning the training data, facilitating the acquisition of adaptive weight information from different blocks within each encoder stage.
Considering the $s$-th stage of encoder that consists of $d$ Transformer blocks, the Gaussian weights $W_{s,l}$ and aggregated features $\mathbf{Z}_s$ can be computed as:
\begin{equation}
\begin{aligned}
    \centering
    W_{s,l} &= \exp\left(-\frac{1}{2}\frac{(d-l+1)^2}{\sigma^2}\right), \quad l \in [1, d],\\
        \mathbf{Z}_s&=\sum\limits_{l=1}^d \mathbf{H}_l \cdot W_{s,l},
    \end{aligned}
    \label{eq4}
\end{equation} 
where the variance parameter $\sigma$ of the Gaussian function is set to 1 by default.
$l$ corresponds to the index of the Transformer block.
Increasing $\sigma$ results in uniform weighting across Transformer blocks while decreasing $\sigma$ leads to the dependence on single block features (as demonstrated in our ablation experiments in~\secref{sec: a-exper} of the Appendix). 
By setting variance parameter $\sigma$, the NGA module can assign a higher weight to nearby blocks and a lower weight to distant ones, facilitating more effective and flexible integration of features across different depth levels.

\section{Experiments}
\subsection{Datasets and Evaluation Metrics}
To evaluate the effectiveness of our proposed method,
we perform extensive experiments on three widely used benchmark datasets, including
COCO-Stuff~\cite{Caesar_2018_CVPR}, Pascal-VOC~\cite{everingham2015pascal}, and Pascal-Context~\cite{mottaghi2014role}.
The split of seen and unseen categories follows the common settings of the previous works~\cite{Zhou_2023_CVPR},
and the mean IoU and harmonic mean IoU of both seen and unseen categories are reported.
Further details on dataset statistics, data splitting, and evaluation metrics can be found in~\secref{A-sec: dataset} of the Appendix.

\begin{table*}[t]
    \centering
    \vskip -0.1in
    \caption{Comparison with the \emph{\textbf{inductive}} and \emph{\textbf{transductive}} state-of-the-art zero-shot segmentation methods on COCO-Stuff 164K, and PASCAL VOC 2012 datasets.
    R denotes ResNet~\cite{he2016deep}.
    ST represents re-train the model with the generating pseudo-labels on unseen classes.
    Our \ourMthd~integrates features extracted from layers 6 to 12 of the image encoder using a three-stage cascade decoder: \{6-8\}, \{9-11\}, \{12\}.
    }
    \label{tab: sota-inductive}
    \begin{center}
    \begin{small}
    \setlength{\tabcolsep}{1.6mm}{
    \begin{tabular}{lccccccccccc}
    \toprule
    \multirow{2}{*}{Methods}& \multirow{2}{*}{Backbone} & \multirow{2}{*}{Segmentor\footnotemark[2]} & \multicolumn{3}{c}{COCO-Stuff 164K~(171)}&\multicolumn{3}{c}{PASCAL VOC 2012~(20)} \\  
    \cmidrule(lr){4-6}  \cmidrule(lr){7-9}
    & &  & mIoU$^S$$\uparrow$& mIoU$^U$$\uparrow$& hIoU$\uparrow$& mIoU$^S$$\uparrow$& mIoU$^U$$\uparrow$& hIoU$\uparrow$ \\
    \midrule
    \multicolumn{9}{l}{\emph{\textbf{Inductive: training images do not contain any unseen objects.}}} \\
    ZegFormer~\cite{ding2022decoupling}&R101\&CLIP-B& MaskFormer &36.6& 33.2& 34.8& 86.4& 63.6& 73.3\\
    Zsseg~\cite{xu2021simple}&R101\&CLIP-B& MaskFormer &39.3& 36.3& 37.8& 83.5& 72.5& 77.5\\
    DeOP~\cite{han2023open}&R101\&CLIP-B& MaskFormer&38.0&38.4&38.2&88.2&74.6&80.8 \\
    Zsseg+MAFT~\cite{jiao2023learning}&R101\&CLIP-B& MaskFormer& 40.6& 40.1&40.3&88.4&66.2&75.7 \\
    SPNet-C~\cite{xian2019semantic} & R101& W2V\&FT& 35.2& 8.7 & 14.0& 78.0& 15.6& 26.1 \\
    ZS3Net~\cite{bucher2019zero}  & R101& W2V    & 34.7& 9.5 & 15.0& 77.3& 17.7& 28.7 \\
    CaGNet~\cite{gu2020context}   & R101& W2V\&FT& 33.5& 12.2& 18.2& 78.4& 26.6& 39.7 \\
    SIGN~\cite{cheng2021sign}   & R101& W2V\&FT     & 32.3& 15.5& 20.9& 75.4& 28.9& 41.7 \\
    JoEm~\cite{baek2021exploiting}   & R101& W2V    & -   &    -&    -& 77.7& 32.5& 45.9 \\
    ZegCLIP~\cite{Zhou_2023_CVPR}&CLIP-B &SegViT &40.2& 41.4& 40.8& 91.9& 77.8& 84.3 \\
    \ourMthd~(Ours)&CLIP-B &SegViT&\textbf{41.1}&\textbf{43.4}&\textbf{42.2}&\textbf{92.7}&\textbf{83.1}&\textbf{87.7} \\
    \midrule
    \midrule 
    \multicolumn{9}{l}{\emph{\textbf{Transductive: training images employ the names of unseen classes.}}} \\
    Zsseg+ST~\cite{xu2021simple}&R101\&CLIP-B& MaskFormer &39.6& 43.6& 41.5& 79.2& 78.1& 79.3\\
    FreeSeg~\cite{Qin_2023_CVPR}&R101\&CLIP-B& Mask2Former& 42.4& 42.2& 42.3&91.9 &78.6&84.7\\
    FreeSeg+MAFT~\cite{jiao2023learning}&R101\&CLIP-B& Mask2Former&\textbf{44.1}&55.2&49.0&90.0&86.3&88.1 \\
    SPNet-C+ST~\cite{xian2019semantic} & R101& W2V\&FT& 34.6& 26.9 & 30.3& 77.8& 25.8& 38.8 \\
    ZS5Net~\cite{bucher2019zero}  & R101& W2V    & 34.9& 10.6 & 16.2& 78.0& 21.2& 33.3 \\
    CaGNet+ST~\cite{gu2020context}   & R101& W2V\&FT& 35.6& 13.4& 19.5& 78.6& 30.3& 43.7 \\
    MaskCLIP$\textbf{\textbf{+}}$~\cite{zhou2022extract}  & R101 & DeepLabv2& 38.1& 54.7& 45.0& 88.8& 86.1& 87.4\\
    MVP-SEG$\textbf{\textbf{+}}$~\cite{guo2023mvp}& R101 & DeepLabv2 &38.3&55.8&39.9& 44.9&67.5&54.0\\
    ZegCLIP+ST~\cite{Zhou_2023_CVPR}&CLIP-B &SegViT &    40.7& 59.9& 48.5& 92.3& 89.9& 91.1 \\
    TagCLIP+ST~\cite{li2023tagclip}&CLIP-B &SegViT&40.4&60.0&48.3&\textbf{94.3}&92.7& \textbf{93.5}\\
    \ourMthd+ST~(Ours) &CLIP-B &SegViT&41.7& \textbf{62.5}& \textbf{50.0}& 93.3& \textbf{93.4} & 93.4 \\
    \bottomrule
    \end{tabular}}
    \end{small}
    \end{center}
    \vskip -0.1in
\end{table*}

\subsection{Implementation Details}
We implement the proposed method on the open-source toolbox MMSegmentation~\cite{mmseg2020} and conduct all experiments using a machine with 4 NVIDIA RTX 3090 GPUs.
%
VIT-B/16~\cite{dosovitskiy2020image}, which contains 12 Transformer blocks, is adopted as the image encoder of CLIP~\cite{pmlr-v139-radford21a}.
The batch size on each GPU is set to 4, and the input image resolution is 512 $\times$ 512.
The optimizer is AdamW~\cite{adamw} with the default training schedule in the MMSeg toolbox.
For a fair comparison,
we use the same number of training iterations on each dataset as ZegCLIP~\cite{Zhou_2023_CVPR}.

\subsection{Comparisons with the State-of-the-art Methods}
To demonstrate the effectiveness of our \ourMthd, 
the evaluation results are compared with previous state-of-the-art methods, 
including two-encoder approaches (\eg ZegFormer~\cite{ding2022decoupling}, Zsseg~\cite{xu2021simple} and DeOP~\cite{han2023open}) 
and one-encoder approaches (\eg ZegCLIP~\cite{Zhou_2023_CVPR}).

\myPara{Comparisons in the inductive setting.}
As shown in~\tabref{tab: sota-inductive}, \ourMthd~remarkably improves the performance in the inductive setting,
where features and annotations for unseen classes are not provided.
It is worth noting that while boosting the results of the seen classes, our method improves the performances of the unseen classes. 
For example, \ourMthd~promotes the state-of-the-art performance by 2.0\% on COCO and 5.3\% on Pascal VOC in terms of mIoU for unseen classes, demonstrating its robust generalization capabilities of zero-shot segmentation.

\begin{table}[t]
    \centering
    \vskip -0.1in
    \caption{Comparison with the state-of-the-art zero-shot segmentation methods on PASCAL Context dataset. 
    }
    \label{tab: context}
    \begin{center}
    \begin{small}
    \setlength{\tabcolsep}{1.0mm}{
    \begin{tabular}{lcccccc}
    \toprule
    \multirow{2}{*}{Methods} & \multicolumn{3}{c}{PASCAL Context~(59)} \\  
    \cmidrule(lr){2-4} 
    & mIoU$^S$$\uparrow$& mIoU$^U$$\uparrow$& hIoU$\uparrow$\\
    \midrule
    \multicolumn{1}{l}{\emph{\textbf{Inductive}}} \\
    SPNet-C~\cite{xian2019semantic} & 27.1& 9.8 & 14.4 \\
    ZS3Net~\cite{bucher2019zero} & 20.8&12.7& 15.8 \\
    CSRL~\cite{li2020consistent} & 29.4& 14.6& 19.5 \\
    JoEm~\cite{baek2021exploiting} &33.0 & 14.9& 20.5 \\
    ZegCLIP~\cite{Zhou_2023_CVPR}&53.8& 45.5 & 49.3 \\
    \ourMthd~(Ours)& \textbf{55.9} & \textbf{47.2} & \textbf{51.2}\\
    \midrule
    \multicolumn{1}{l}{\emph{\textbf{Transductive}}} \\
    ZS5Net~\cite{bucher2019zero}& 27.0& 20.7& 23.4 \\
    ZegCLIP+ST~\cite{Zhou_2023_CVPR}& 54.5 & 41.4 & 47.1\\
    \ourMthd+ST~(Ours)& \textbf{56.4}& \textbf{55.0}& \textbf{55.7}\\
    \bottomrule
    \end{tabular}}
    \end{small}
    \end{center}
    \vskip -0.20in
\end{table}

\footnotetext[2]{The description of segmentors is acquired from~\cite{zhu2023survey}. MaskFormer and MasksFormer are proposed by~\cite{cheng2021per} and~\cite{cheng2022masked}; DeepLabv2 is proposed by~\cite{chen2018encoder}; W2V is proposed by~\cite{mikolov2013distributed}; SegViT is proposed by~\cite{zhang2022segvit}.}

\myPara{Comparisons in the transductive setting.}
We further evaluate the transferability of \ourMthd~in the transductive setting, where models are retrained by generating pseudo labels for unseen pixels and utilizing ground truth labels for seen pixels.
\tabref{tab: sota-inductive} demonstrates that our model significantly improves the performance for unseen classes while consistently maintaining excellent performance across seen classes after transductive self-training.

To further validate the effectiveness of our \ourMthd, we conduct comparisons with other methods on the PASCAL Context dataset.
As shown in~\tabref{tab: context}, our \ourMthd~consistently outperforms other methods, particularly regarding mIoU for unseen classes.
The results above clearly demonstrate the effectiveness of our proposed methods.
For additional experimental results on the effectiveness and generality of our method, please refer to~\secref{exp: extend}.

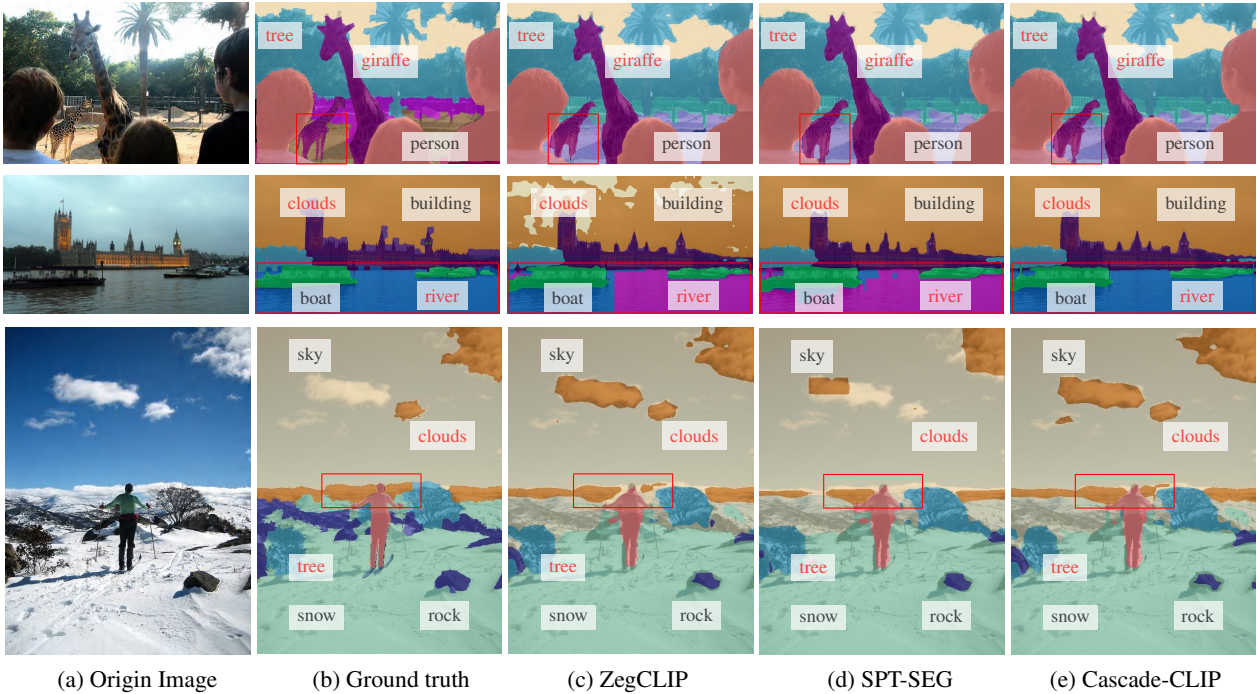
\begin{figure*}[t]
\centering
    \vspace{1pt}
    \begin{minipage}[t]{0.19\linewidth}
        \centering
        \begin{overpic}[width=1\textwidth]{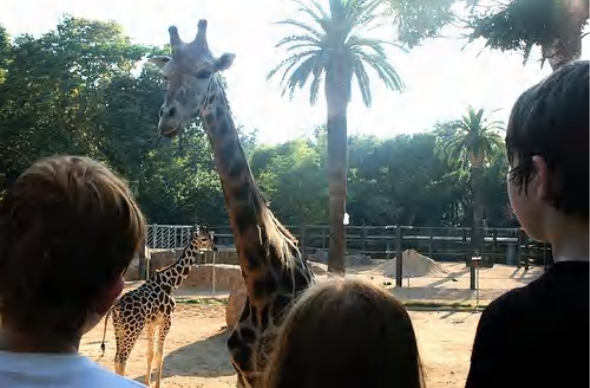}
        \end{overpic}
    \end{minipage}
    \begin{minipage}[t]{0.19\linewidth}
    \centering
    \begin{overpic}[width=1\textwidth]{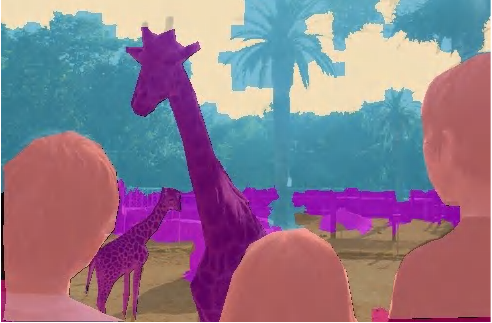}
        \put(1,50){\transparent{0.7}\colorbox{white}{\textcolor{red}{\scriptsize tree}}}
        \put(60,6){\transparent{0.7}\colorbox{white}{\textcolor{black}{\scriptsize person}}}    
        \put(40,40){\transparent{0.7}\colorbox{white}{\textcolor{red}{\scriptsize giraffe}}}
        \put(17,0.1){\color{red}\framebox(20,20){}}
    \end{overpic}
    \end{minipage}   
    \begin{minipage}[t]{0.19\linewidth}
        \centering
        \begin{overpic}[width=1\textwidth]{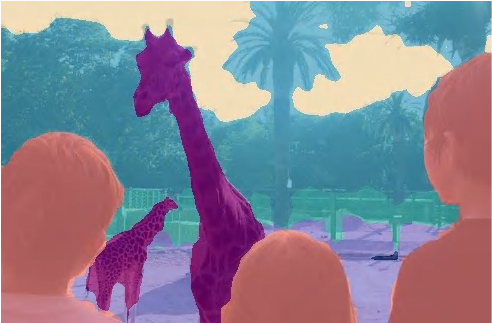}
            \put(1,50){\transparent{0.7}\colorbox{white}{\textcolor{red}{\scriptsize tree}}}
            \put(60,6){\transparent{0.7}\colorbox{white}{\textcolor{black}{\scriptsize person}}}    
            \put(40,40){\transparent{0.7}\colorbox{white}{\textcolor{red}{\scriptsize giraffe}}}
            \put(17,0.1){\color{red}\framebox(20,20){}}    
        \end{overpic}
    \end{minipage}
    \begin{minipage}[t]{0.19\linewidth}
        \centering
        \begin{overpic}[width=1\textwidth]{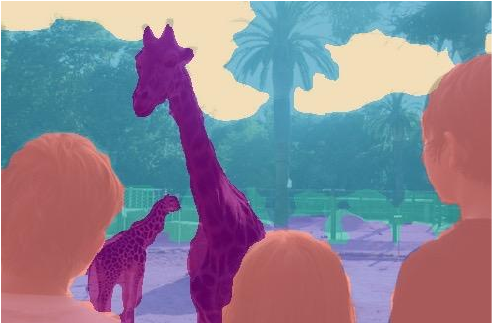}
            \put(1,50){\transparent{0.7}\colorbox{white}{\textcolor{red}{\scriptsize tree}}}
            \put(60,6){\transparent{0.7}\colorbox{white}{\textcolor{black}{\scriptsize person}}}    
            \put(40,40){\transparent{0.7}\colorbox{white}{\textcolor{red}{\scriptsize giraffe}}}
            \put(17,0.1){\color{red}\framebox(20,20){}}
        \end{overpic}
    \end{minipage}   
    \begin{minipage}[t]{0.19\linewidth}
    \centering
    \begin{overpic}[width=1\textwidth]{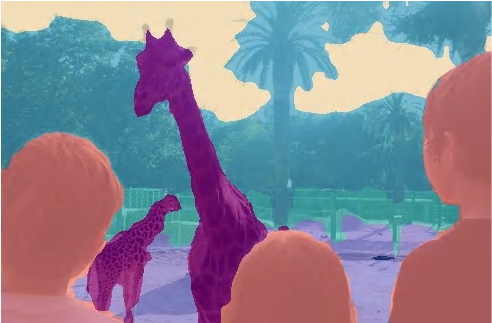}
        \put(1,50){\transparent{0.7}\colorbox{white}{\textcolor{red}{\scriptsize tree}}}
        \put(60,6){\transparent{0.7}\colorbox{white}{\textcolor{black}{\scriptsize person}}}    
        \put(40,40){\transparent{0.7}\colorbox{white}{\textcolor{red}{\scriptsize giraffe}}}
        \put(17,0.1){\color{red}\framebox(20,20){}}    
    \end{overpic}
    \end{minipage} 
    \vspace{4pt}
    \\
    \begin{minipage}[t]{0.19\linewidth}
        \centering
        \begin{overpic}[width=1\textwidth]{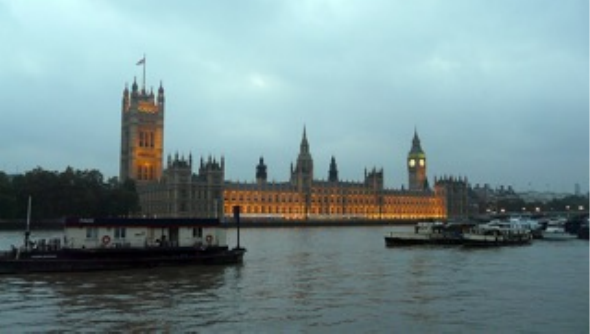}
        \end{overpic}
    \end{minipage}
    \begin{minipage}[t]{0.19\linewidth}
    \centering
    \begin{overpic}[width=1\textwidth]{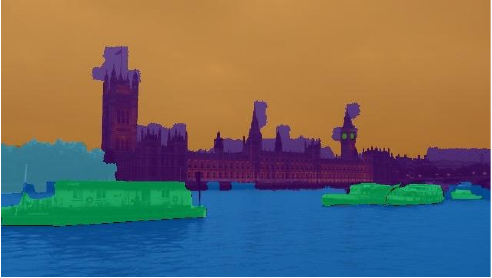}
        \put(10,43){\transparent{0.7}\colorbox{white}{\textcolor{red}{\scriptsize clouds}}}
        \put(66,5){\transparent{0.7}\colorbox{white}{\textcolor{red}{\scriptsize river}}}
        \put(60,43){\transparent{0.7}\colorbox{white}{\textcolor{black}{\scriptsize building}}}    
        \put(15,4){\transparent{0.7}\colorbox{white}{\textcolor{black}{\scriptsize boat}}}    
        \put(1.0,0.6){\color{red}\framebox(98,20){}}
    \end{overpic}
    \end{minipage}   
    \begin{minipage}[t]{0.19\linewidth}
        \centering
        \begin{overpic}[width=1\textwidth]{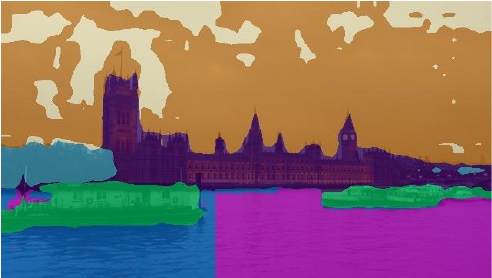}
            \put(10,43){\transparent{0.7}\colorbox{white}{\textcolor{red}{\scriptsize clouds}}}
            \put(66,5){\transparent{0.7}\colorbox{white}{\textcolor{red}{\scriptsize river}}}
            \put(60,43){\transparent{0.7}\colorbox{white}{\textcolor{black}{\scriptsize building}}}    
            \put(15,4){\transparent{0.7}\colorbox{white}{\textcolor{black}{\scriptsize boat}}}        
            \put(1.0,0.6){\color{red}\framebox(98,20){}}       
        \end{overpic}
    \end{minipage} 
    \begin{minipage}[t]{0.19\linewidth}
        \centering
        \begin{overpic}[width=1\textwidth]{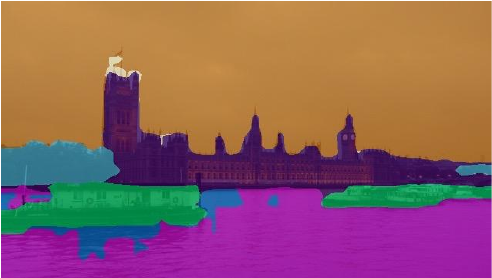}
            \put(10,43){\transparent{0.7}\colorbox{white}{\textcolor{red}{\scriptsize clouds}}}
            \put(66,5){\transparent{0.7}\colorbox{white}{\textcolor{red}{\scriptsize river}}}
            \put(60,43){\transparent{0.7}\colorbox{white}{\textcolor{black}{\scriptsize building}}}    
            \put(15,4){\transparent{0.7}\colorbox{white}{\textcolor{black}{\scriptsize boat}}}        
            \put(1.0,0.6){\color{red}\framebox(98,20){}}        
        \end{overpic}
    \end{minipage} 
    \begin{minipage}[t]{0.19\linewidth}
        \centering
        \begin{overpic}[width=1\textwidth]{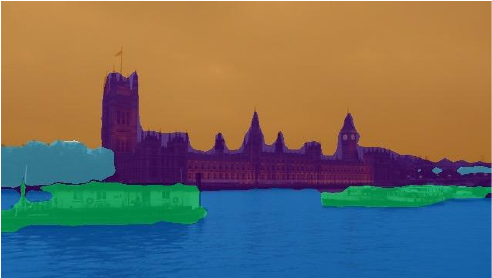}
            \put(10,43){\transparent{0.7}\colorbox{white}{\textcolor{red}{\scriptsize clouds}}}
            \put(66,5){\transparent{0.7}\colorbox{white}{\textcolor{red}{\scriptsize river}}}
            \put(60,43){\transparent{0.7}\colorbox{white}{\textcolor{black}{\scriptsize building}}}    
            \put(15,4){\transparent{0.7}\colorbox{white}{\textcolor{black}{\scriptsize boat}}}        
            \put(1.0,0.6){\color{red}\framebox(98,20){}}
        \end{overpic}
    \end{minipage} 
    \vspace{4pt}
    \\
    \begin{minipage}[t]{0.19\linewidth}
        \centering
        \begin{overpic}[width=1\textwidth]{visualisation_transductive4-1}
        \put(16,-10){{\footnotesize (a) Origin Image}}
        \end{overpic}
    \end{minipage}
    \begin{minipage}[t]{0.19\linewidth}
        \centering
        \begin{overpic}[width=1\textwidth]{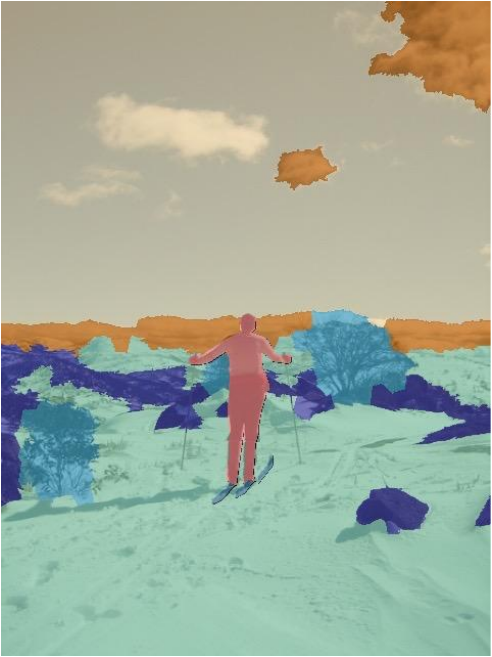}
        \put(10,90){\transparent{0.7}\colorbox{white}{\textcolor{black}{\scriptsize sky}}}    
        \put(10,10){\transparent{0.7}\colorbox{white}{\textcolor{black}{\scriptsize snow}}}    
        \put(47,65){\transparent{0.7}\colorbox{white}{\textcolor{red}{\scriptsize clouds}}}
        \put(10,25){\transparent{0.7}\colorbox{white}{\textcolor{red}{\scriptsize tree}}}
        \put(50,10){\transparent{0.7}\colorbox{white}{\textcolor{black}{\scriptsize rock}}}    
        \put(20,45){\color{red}\framebox(30,10){}}
        \put(17,-10){{\footnotesize (b) Ground truth}}
    \end{overpic}
    \end{minipage}
    \begin{minipage}[t]{0.19\linewidth}
        \centering
        \begin{overpic}[width=1\textwidth]{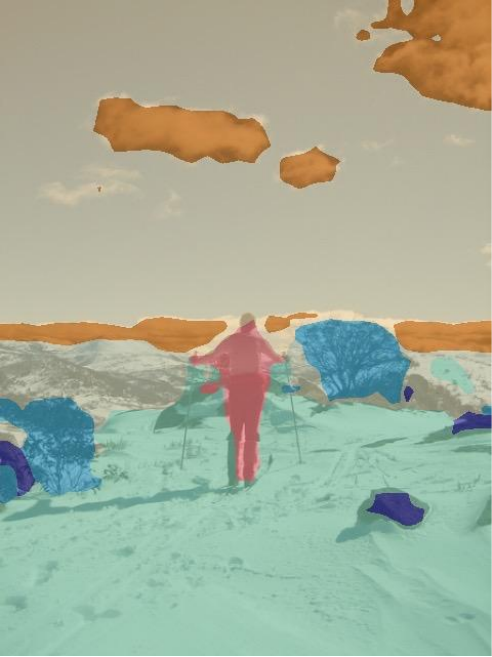}
        \put(10,90){\transparent{0.7}\colorbox{white}{\textcolor{black}{\scriptsize sky}}}    
        \put(10,10){\transparent{0.7}\colorbox{white}{\textcolor{black}{\scriptsize snow}}}    
        \put(47,65){\transparent{0.7}\colorbox{white}{\textcolor{red}{\scriptsize clouds}}}
        \put(10,25){\transparent{0.7}\colorbox{white}{\textcolor{red}{\scriptsize tree}}}
        \put(50,10){\transparent{0.7}\colorbox{white}{\textcolor{black}{\scriptsize rock}}}    
        \put(20,45){\color{red}\framebox(30,10){}}
        \put(18,-10){{\footnotesize (c) ZegCLIP}}
        \end{overpic}
    \end{minipage}
    \begin{minipage}[t]{0.19\linewidth}
        \centering
        \begin{overpic}[width=1\textwidth]{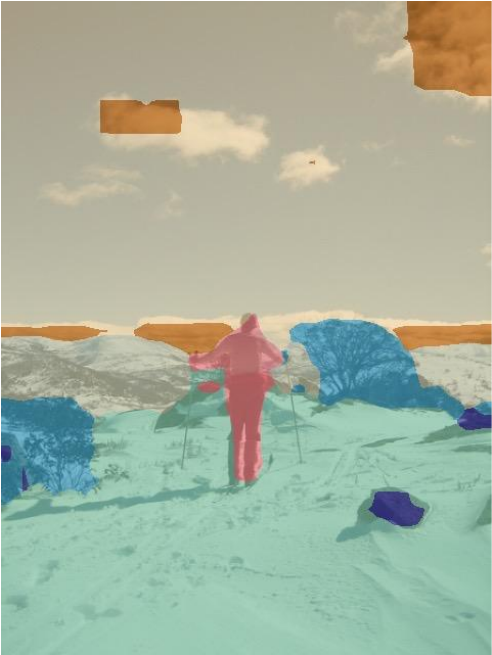}
        \put(10,90){\transparent{0.7}\colorbox{white}{\textcolor{black}{\scriptsize sky}}}    
        \put(10,10){\transparent{0.7}\colorbox{white}{\textcolor{black}{\scriptsize snow}}}    
        \put(47,65){\transparent{0.7}\colorbox{white}{\textcolor{red}{\scriptsize clouds}}}
        \put(10,25){\transparent{0.7}\colorbox{white}{\textcolor{red}{\scriptsize tree}}}
        \put(50,10){\transparent{0.7}\colorbox{white}{\textcolor{black}{\scriptsize rock}}}    
        \put(20,45){\color{red}\framebox(30,10){}}
        \put(21,-10){{\footnotesize (d) SPT-SEG}}
        \end{overpic}
    \end{minipage}
    \begin{minipage}[t]{0.19\linewidth}
        \centering
        \begin{overpic}[width=1\textwidth]{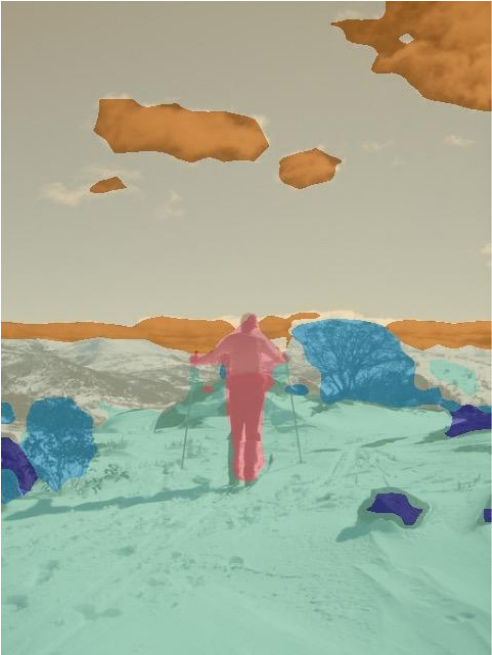}
        \put(10,90){\transparent{0.7}\colorbox{white}{\textcolor{black}{\scriptsize sky}}}    
        \put(10,10){\transparent{0.7}\colorbox{white}{\textcolor{black}{\scriptsize snow}}}    
        \put(47,65){\transparent{0.7}\colorbox{white}{\textcolor{red}{\scriptsize clouds}}}
        \put(10,25){\transparent{0.7}\colorbox{white}{\textcolor{red}{\scriptsize tree}}}
        \put(50,10){\transparent{0.7}\colorbox{white}{\textcolor{black}{\scriptsize rock}}}    
        \put(20,45){\color{red}\framebox(30,10){}}
        \put(12,-10){{\footnotesize (e) \ourMthd}}
        \end{overpic}
    \end{minipage}
    \vspace{5pt}
\caption{Qualitative transductive results on COCO-Stuff 164K. 
The \textcolor{black}{black} and \textcolor{red}{red} tags represent seen and unseen classes, respectively.  
}
\label{visualisation_transductive}
\vskip -0.15in
\end{figure*}

\myPara{Qualitative results.}
\figref{visualisation_transductive} shows the segmentation results of 
the baseline and our proposed \ourMthd~on seen and unseen classes. 
\ourMthd~shows impressive segmentation ability for both seen and unseen classes 
and can clearly distinguish similar unseen classes.
For example, our method can better distinguish the `giraffe' regions from the `tree' regions (\figref{visualisation_transductive}(1)),
the `boat' regions from the `river' regions (\figref{visualisation_transductive}(2)),
and the `clouds' regions from the `sky' regions (\figref{visualisation_transductive}(3)).
More qualitative results are available in~\secref{A-sec: visualization} of the Appendix.

\subsection{Ablation Study}
\label{exp: ablation}
\myPara{Component-wise ablations.}
To understand the effect of each component in our \ourMthd, including the cascaded decoders and the NGA modules,
we initiate with the baseline ZegCLIP, which employs the visual features of CLIP's last layer, and then gradually incorporate each proposed module.
As shown in~\tabref{tab: component-wise},
employing cascaded decoders can capture distinct and complementary information from different blocks in the encoder,
improving 3.1\% mIoU scores on unseen classes (the 2nd result). 
On this basis, the NGA module is introduced to aggregate rich local information of objects within each split encoder, 
further enhancing the mIoU scores on unseen classes (the 3rd result).

\begin{table}[t]
    \centering
    \caption{Ablation on components of \ourMthd.}
    \label{tab: component-wise}
    \begin{center}
    \begin{small}
    \setlength{\tabcolsep}{1.5mm}{
    \begin{tabular}{ccccc}
    \toprule
    Cascaded decoders& NGA & mIoU$^S$$\uparrow$& mIoU$^U$$\uparrow$& hIoU$\uparrow$ \\    
    \midrule
    \xmark& \xmark & 40.1 & 39.5 & 39.8 \\
    \cmark& \xmark & 40.8 & 42.6 & 41.7 \\
    \cmark& \cmark & \textbf{41.1} & \textbf{43.4} & \textbf{42.2} \\
    \bottomrule
    \end{tabular}}
    \end{small}
    \end{center}
    \vskip -0.2in
\end{table}

\begin{table}[t]
    \centering
    \small
    \caption{Effect of different block splitting manners.
    The elements in \{\} means the block numbers that are used to fuse features in the encoder. 
    When the number of blocks exceeds 1, an NGA is used.
    }
    \label{tab: effect-part-decouple}
    \setlength{\tabcolsep}{0.8mm}
    \begin{tabular}{lcccc} \toprule
     Block splitting manner & \#Decoders & mIoU$^S$$\uparrow$& mIoU$^U$$\uparrow$& hIoU$\uparrow$ \\
    \midrule
    \{4-6\}, \{7-9\}, \{10-12\}& 3 & 40.4& 42.7 &41.5\\
     \{6-8\}, \{9-10\}, \{11-12\}& 3 & 40.7& 42.4 &41.5\\
    \{6-8\}, \{9-11\}, \{12\} & 3 & \textbf{41.1} & \textbf{43.4} & \textbf{42.2} \\
    \bottomrule
    \end{tabular}
    \vskip -0.1in
\end{table}

\myPara{Effect of the proposed block splitting manner.}
The cascaded decoder architecture is pivotal in our \ourMthd{} due to its ability to preserve visual-language associations.
Our analysis in \tabref{tab: effect-part-decouple} indicates that separating the last block into an independent stage (the 3rd result) is more effective than other combinations of partitioning strategies (the 1st result and the 2nd result).
This is because the last layer feature of CLIP's image encoder possesses the strongest association with text embedding, and matching it to a separate decoder reduces the destruction of this correlation. 


\begin{table}[!t]
    \centering
    \caption{Numbers of cascaded decoders and corresponding blocks in the visual encoder of \ourMthd.
    %
    }
    \label{tab: num-seg}
    \begin{center}
    \begin{small}
    \setlength{\tabcolsep}{0.6mm}{
    \begin{tabular}{lcccc}
    \toprule
    Block splitting manner  & Number & mIoU$^S$$\uparrow$& mIoU$^U$$\uparrow$& hIoU$\uparrow$ \\   
    \midrule
    \multicolumn{5}{l}{Number of cascaded decoders}\\
    \{12\} (Baseline) &1 & 40.1 & 39.5 & 39.8 \\
    \{9-11\}, \{12\} & 2 & 40.2 & 40.2 & 40.2 \\
    \{6-8\}, \{9-11\}, \{12\} & 3 & \textbf{41.1} & \textbf{43.4} & \textbf{42.2} \\
    \{3-5\}, \{6-8\}, \{9-11\}, \{12\} & 4 & 40.8 & 42.5 & 41.7 \\
    \midrule
    \multicolumn{5}{l}{Number of blocks}\\
    \{8-9\}, \{10-11\}, \{12\} & 2 & 40.7 & 41.3 & 41.0 \\
    \{6-8\}, \{9-11\}, \{12\} & 3 & \textbf{41.1} & \textbf{43.4} & \textbf{42.2} \\
    \{4-7\}, \{8-11\}, \{12\} & 4 & 41.1 & 41.5 & 41.3 \\
    \bottomrule
    \end{tabular}}
    \end{small}
    \end{center}
    \vskip -0.2in
\end{table}
\begin{table}[!t]
    \centering
    \caption{\ourMthd~vs. other methods with multi-decoders.
     Utilizing the last-layer features or directly fusing multi-layer features with the same blocks as our \ourMthd~and aligning them with multiple decoders results in a decline in performance.
    }
    \label{tab: multi-decoders}
    \begin{center}
    \begin{small}
    \setlength{\tabcolsep}{1mm}{
    \begin{tabular}{lcccc}
    \toprule
    Description & Param. (M)& mIoU$^S$$\uparrow$& mIoU$^U$$\uparrow$& hIoU$\uparrow$ \\    
    \midrule
    Last-layer&40.5 & 40.3& 40.6 &40.5\\
    Multi-layer fusion&40.5 & 39.9& 35.7& 37.7\\
    \ourMthd&40.5 & \textbf{41.1} & \textbf{43.4} & \textbf{42.2} \\
    \bottomrule
    \end{tabular}}
    \end{small}
    \end{center}
    \vskip -0.1in
\end{table}
\begin{table}[!t]
    \centering
    \vskip -0.1in
    \caption{Comparison of different aggregation methods.
    $\dagger$ denotes that the weights are trainable.}
    \label{tab: NGA-others}
    \begin{center}
    \begin{small}
    \setlength{\tabcolsep}{3.8mm}{
    \begin{tabular}{lccc}
    \toprule
    Description & mIoU$^S$$\uparrow$& mIoU$^U$$\uparrow$& hIoU$\uparrow$ \\    
    \midrule
    Concat & 39.9&41.0&40.5\\
    Self-attention& 37.7&39.0&38.4\\
    Sum & 39.3& 42.0& 40.6\\
    NGA &40.9 & 43.0 & 41.9 \\
    \midrule
    Sum$\dagger$& 39.9&41.2&40.5\\
    NGA$\dagger$& \textbf{41.1} & \textbf{43.4} & \textbf{42.2} \\
    \bottomrule
    \end{tabular}}
    \end{small}
    \end{center}
    \vskip -0.1in
\end{table}

\myPara{Number of cascaded decoders and number of blocks in each stage.}
To show the importance of information fusion across different layers, we present the performance of \ourMthd~with different numbers of cascaded decoders and different blocks in each corresponding encoder stage in~\tabref{tab: num-seg}.
We can see that increasing the number of cascaded decoders from 1 to 3 gradually improves the segmentation performance.
This indicates the complementarity of features from various layers compared to previous works only using the features from the last layer.
Our default value of Transformer blocks per stage is 3.
Reducing the number of blocks to 2 causes a degradation in performance due to the neglect of mid-layer features.
The best performance is achieved by cascading three decoders (including an extra decoder for the last block).
Note that we do not use the beginning blocks as they encode features with little semantics.

To demonstrate the effectiveness of our designs in leveraging CLIP's multi-level features, we also present cosine similarity maps of features and qualitative segmentation results.
At the top of \figref{A-visualisation_inductive}, we show the patch similarity of unseen classes not included in the training process.
We observe that the intermediate layers of our method contain detailed information on local objects, including boundaries.
Moreover, as illustrated at the bottom of~\figref{A-visualisation_inductive}, by leveraging these distinctive features, our \ourMthd~improves segmentation performance for both seen and unseen classes compared to using only last-block features.

\begin{table}[!t]
    \centering
    \caption{Effect of independent/shared text embedding.
    %
    }
    \label{tab: indep-shated-text}
    \begin{center}
    \begin{small}
    \setlength{\tabcolsep}{4.2mm}{
    \begin{tabular}{lcccc}
    \toprule
    Description & mIoU$^S$$\uparrow$& mIoU$^U$$\uparrow$& hIoU$\uparrow$ \\    
    \midrule
    Shared & 40.5& 42.4 & 41.4\\
    Independent & \textbf{41.1} & \textbf{43.4} & \textbf{42.2} \\
    \bottomrule
    \end{tabular}}
    \end{small}
    \end{center}
    \vskip -0.1in
\end{table}
\begin{table}[t]
    \centering
    \caption{Extending \ourMthd~to existing methods to improve zero-shot segmentation results.}
    \label{tab: our-othermethods}
    \begin{center}
    \begin{small}
    \setlength{\tabcolsep}{0.9mm}{
    \begin{tabular}{lccccccc}
    \toprule
    \multirow{2}{*}{Methods} & \multicolumn{2}{c}{COCO-Stuff 164K}&\multicolumn{2}{c}{VOC 2012} \\  
    \cmidrule(lr){2-3}  \cmidrule(lr){4-5}
    & mIoU$^S$$\uparrow$& mIoU$^U$$\uparrow$& mIoU$^S$$\uparrow$& mIoU$^U$$\uparrow$& \\
    \midrule
    \multicolumn{5}{l}{\emph{\textbf{Inductive}}} \\
    Frozen CLIP & 32.3  &32.5 & 85.9&59.5 \\
    Frozen CLIP+Ours &\textbf{36.3}&\textbf{35.3}&\textbf{89.0}&\textbf{69.7}\\
    \midrule 
    ZegCLIP &    40.2& 41.4& 91.9& 77.8\\
    ZegCLIP+Ours &\textbf{41.1}&\textbf{43.4}&\textbf{92.7}&\textbf{83.1}\\
    \midrule 
    SPT-SEG &  38.0  &40.7& 92.0& 85.0 \\
    SPT-SEG+Ours &\textbf{40.2}&\textbf{43.6}  &\textbf{92.1}&\textbf{86.1}\\
    \midrule 
    \midrule 
    \multicolumn{5}{l}{\emph{\textbf{Transductive}}} \\
    ZegCLIP+ST& 40.7& 59.9& 92.3& 89.9\\
    ZegCLIP+Ours+ST&\textbf{41.7}& \textbf{62.5}& \textbf{93.3}& \textbf{93.4} \\
    \midrule 
    SPT-SEG& 40.4&57.5 & 93.6& 92.2\\
    SPT-SEG+Ours+ST&\textbf{41.7}&\textbf{62.1}&\textbf{93.8}&\textbf{94.4}  \\
    \midrule
    \midrule 
    \multicolumn{5}{l}{\emph{\textbf{Fully Supervised}}} \\
    ZegCLIP &    40.7& 63.2& 92.4& 90.9\\
    ZegCLIP+Ours &\textbf{41.5}&\textbf{64.0}&\textbf{93.7}&\textbf{94.6}\\
    \bottomrule
    \end{tabular}}
    \end{small}
    \end{center}
    \vskip -0.2in
\end{table}
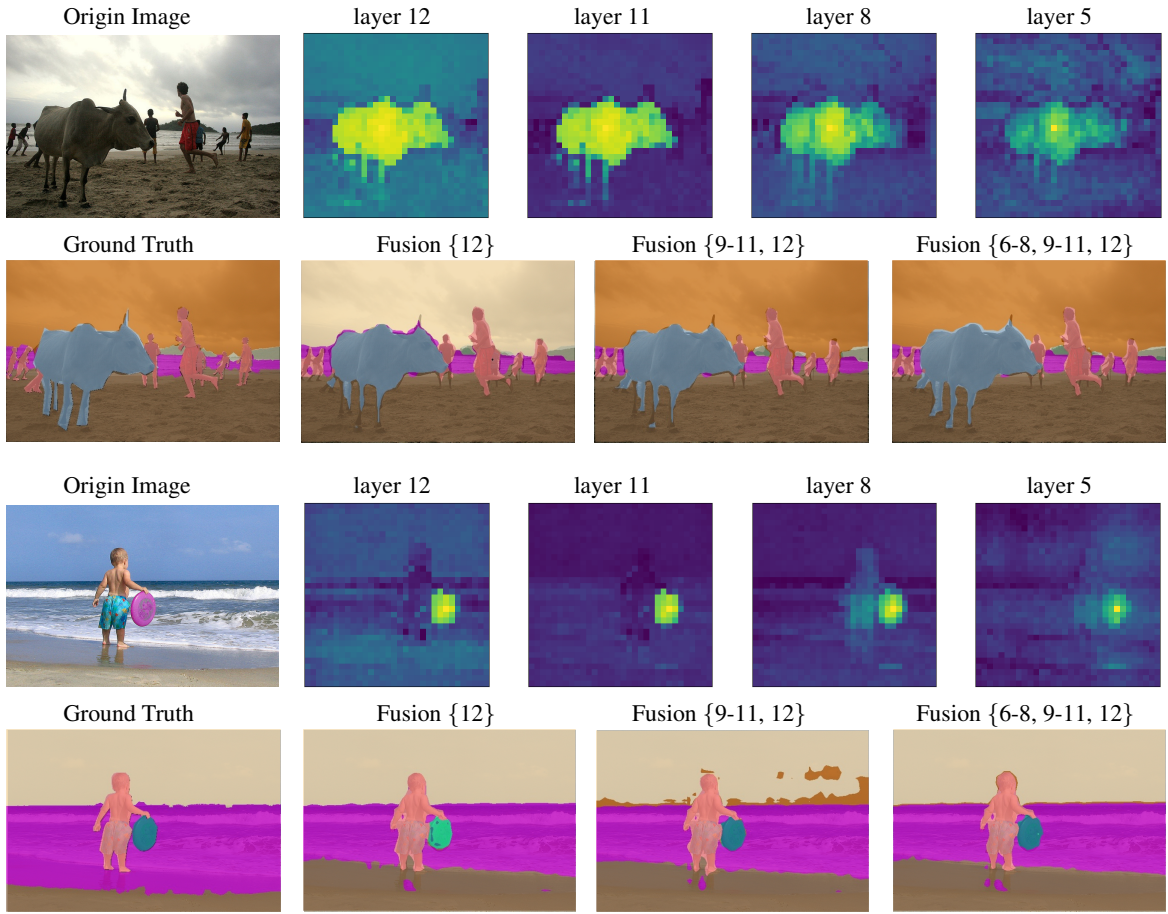
\begin{figure*}[t]
\centering
    \vspace{20pt}
    \begin{minipage}[t]{0.9\linewidth}
        \centering
        \begin{overpic}[width=1\textwidth]{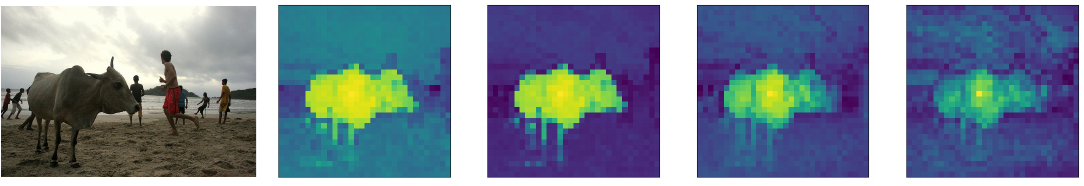}
        \put(5,17.5){\footnotesize{Origin Image}}
        \put(30,17.5){\footnotesize{layer 12}}
        \put(49,17.5){\footnotesize{layer 11}}
        \put(69,17.5){\footnotesize{layer 8}}
        \put(88,17.5){\footnotesize{layer 5}}
        \end{overpic}
    \end{minipage}
    \vspace{15pt}
    \\
    \begin{minipage}[t]{0.9\linewidth}
        \centering
        \begin{overpic}[width=1\textwidth]{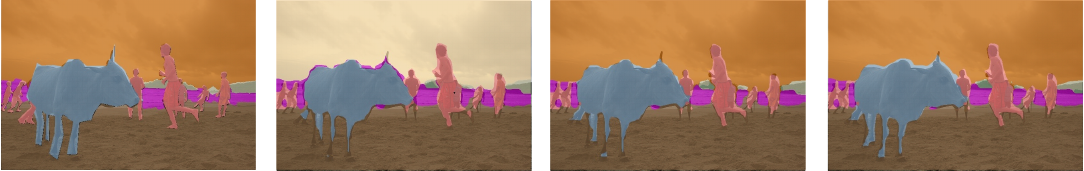}
        \put(5,16.5){\footnotesize{Ground Truth}}
        \put(32,16.5){\footnotesize{Fusion \{12\}}}
        \put(54,16.5){\footnotesize{Fusion \{9-11, 12\}}}
        \put(78.5,16.5){\footnotesize{Fusion \{6-8, 9-11, 12\}}}
        \end{overpic}
    \end{minipage}
        \vspace{20pt}
    \\
    \begin{minipage}[t]{0.9\linewidth}
        \centering
        \begin{overpic}[width=1\textwidth]{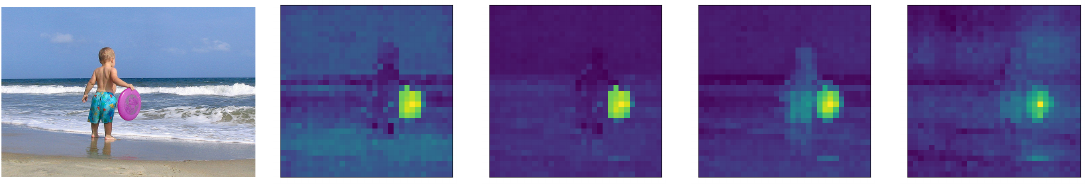}
        \put(5,17.5){\footnotesize{Origin Image}}
        \put(30,17.5){\footnotesize{layer 12}}
        \put(49,17.5){\footnotesize{layer 11}}
        \put(69,17.5){\footnotesize{layer 8}}
        \put(88,17.5){\footnotesize{layer 5}}
        \end{overpic}
    \end{minipage}
    \vspace{15pt}
    \\
    \begin{minipage}[t]{0.9\linewidth}
        \centering
        \begin{overpic}[width=1\textwidth]{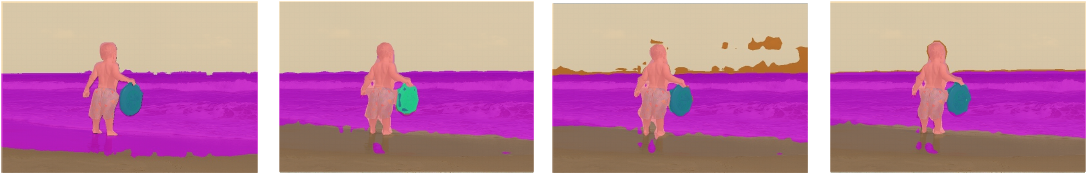}
        \put(5,16.5){\footnotesize{Ground Truth}}
        \put(32,16.5){\footnotesize{Fusion \{12\}}}
        \put(54,16.5){\footnotesize{Fusion \{9-11, 12\}}}
        \put(78.5,16.5){\footnotesize{Fusion \{6-8, 9-11, 12\}}}
        \end{overpic}
    \end{minipage}
\caption{Visualizations of cosine similarity maps about features and qualitative segmentation results.
We visualize feature correspondence through cosine similarity calculations on visual patches of \textbf{unseen} classes from both deep and shallow layers in our \ourMthd.
The inductive segmentation results are generated by cascading different decoders.
}
\label{A-visualisation_inductive}
\end{figure*}

\myPara{NGA v.s. other aggregation methods.}
In each split encoder stage, the differences between features across various layers will disrupt the feature space after aggregating various layers. 
To overcome this issue, we propose Neighborhood Gaussian Aggregation~(NGA) to reduce disruptions in the original feature space by considering the distance between blocks.
As shown in ~\tabref{tab: NGA-others}, our NGA outperforms common aggregation strategies (\eg Sum, Connect and Self-attention).
With learnable weights, our NGA further boosts performance.
This indicates that our NGA, which assigns smaller weights to distant features when fusing multi-level features, is advantageous in improving zero-shot segmentation compared to other feature aggregation methods.

\myPara{\ourMthd~vs. other methods with multi-decoder.}
To demonstrate the efficacy of integrating diverse semantic masks generated from distinct cascaded decoders, rather than introducing extra parameters that could impact performance,
we construct a multi-decoder model based on the last-layer features or the fusion of multi-layer features.
As shown in~\tabref{tab: multi-decoders}, \ourMthd~outperforms the 
last-layer and multi-layer approaches with equivalent parameter amounts.
This indicates that relying only on the last-layer features fails to yield complementary and enhanced segmentation results.
Moreover, the direct fusion of features leads to a decrease in zero-shot capability, which is not improved even with the use of multiple decoders.

\myPara{Effect of independent/shared text embedding.}
Since the features from different split encoder stages exhibit significant discrepancies, it is essential to align distinct text embeddings with the features of each stage.
This is validated by the results presented in~\tabref{tab: indep-shated-text}, where our \ourMthd~with independent text embedding achieves higher mIoU scores than those with shared text embedding. 

\subsection{Extending \ourMthd~to Other Methods}
\label{exp: extend}
Our approach is a generalized framework for improving the zero-shot segmentation capabilities.
Specifically, we can seamlessly integrate \ourMthd~into existing popular zero-shot semantic segmentation methods, \eg Frozen CLIP~\cite{pmlr-v139-radford21a}, ZegCLIP~\cite{Zhou_2023_CVPR} and SPT-SEG~\cite{xu2023spectral}. 
As shown in~\tabref{tab: our-othermethods}, 
our method can significantly enhance these methods' results, proving the proposed approach's generalization ability. 

\section{Conclusions}
This paper focuses on leveraging the intermediate features from CLIP with rich local details but significant differences from deep features to enhance zero-shot semantic segmentation.
By introducing the cascaded mask mechanism, we present the \ourMthd~framework, which aims to effectively align multi-level visual features with the text embeddings in a cascaded way, thereby enhancing CLIP's adaptability from image to pixel level.
Experiments demonstrate the effectiveness of the proposed method.

\subsection*{Impact Statement}

Our work explores how to leverage multi-level features from the pre-trained CLIP to enhance the model's zero-shot capability, which has been proven effective in downstream tasks.
%
Since our method is not designed for a specific application, it does not directly involve societal issues.

\subsection*{Acknowledgments}
This research was supported by NSFC (NO. 62225604, No. 62276145), the Fundamental Research Funds for the Central Universities (Nankai University, 070-63223049), CAST through
Young Elite Scientist Sponsorship Program (No. YESS20210377). Computations were supported by the Supercomputing Center of Nankai University (NKSC).

\bibliography{reference}

\begin{thebibliography}{60}
\providecommand{\natexlab}[1]{#1}
\providecommand{\url}[1]{\texttt{#1}}
\expandafter\ifx\csname urlstyle\endcsname\relax
  \providecommand{\doi}[1]{doi: #1}\else
  \providecommand{\doi}{doi: \begingroup \urlstyle{rm}\Url}\fi

\bibitem[Baek et~al.(2021)Baek, Oh, and Ham]{baek2021exploiting}
Baek, D., Oh, Y., and Ham, B.
\newblock Exploiting a joint embedding space for generalized zero-shot semantic
  segmentation.
\newblock In \emph{ICCV}, pp.\  9536--9545, 2021.

\bibitem[Bucher et~al.(2019)Bucher, Vu, Cord, and P{\'{e}}rez]{bucher2019zero}
Bucher, M., Vu, T., Cord, M., and P{\'{e}}rez, P.
\newblock Zero-shot semantic segmentation.
\newblock In \emph{NeurIPS}, pp.\  466--477, 2019.

\bibitem[Caesar et~al.(2018)Caesar, Uijlings, and Ferrari]{Caesar_2018_CVPR}
Caesar, H., Uijlings, J., and Ferrari, V.
\newblock Coco-stuff: Thing and stuff classes in context.
\newblock In \emph{CVPR}, pp.\  1209--1218, 2018.

\bibitem[Cha et~al.(2023)Cha, Mun, and Roh]{Cha_2023_CVPR}
Cha, J., Mun, J., and Roh, B.
\newblock Learning to generate text-grounded mask for open-world semantic
  segmentation from only image-text pairs.
\newblock In \emph{CVPR}, pp.\  11165--11174, 2023.

\bibitem[Chen et~al.(2018)Chen, Zhu, Papandreou, Schroff, and
  Adam]{chen2018encoder}
Chen, L.-C., Zhu, Y., Papandreou, G., Schroff, F., and Adam, H.
\newblock Encoder-decoder with atrous separable convolution for semantic image
  segmentation.
\newblock In \emph{ECCV}, pp.\  801--818, 2018.

\bibitem[Chen et~al.(2021)Chen, Yuan, Zeng, and Wang]{Chen_2021_CVPR}
Chen, X., Yuan, Y., Zeng, G., and Wang, J.
\newblock Semi-supervised semantic segmentation with cross pseudo supervision.
\newblock In \emph{CVPR}, pp.\  2613--2622, 2021.

\bibitem[Cheng et~al.(2021{\natexlab{a}})Cheng, Schwing, and
  Kirillov]{cheng2021per}
Cheng, B., Schwing, A., and Kirillov, A.
\newblock Per-pixel classification is not all you need for semantic
  segmentation.
\newblock In \emph{NeurIPS}, pp.\  17864--17875, 2021{\natexlab{a}}.

\bibitem[Cheng et~al.(2022{\natexlab{a}})Cheng, Misra, Schwing, Kirillov, and
  Girdhar]{Cheng_2022_CVPR}
Cheng, B., Misra, I., Schwing, A.~G., Kirillov, A., and Girdhar, R.
\newblock Masked-attention mask transformer for universal image segmentation.
\newblock In \emph{CVPR}, pp.\  1290--1299, 2022{\natexlab{a}}.

\bibitem[Cheng et~al.(2022{\natexlab{b}})Cheng, Misra, Schwing, Kirillov, and
  Girdhar]{cheng2022masked}
Cheng, B., Misra, I., Schwing, A.~G., Kirillov, A., and Girdhar, R.
\newblock Masked-attention mask transformer for universal image segmentation.
\newblock In \emph{CVPR}, pp.\  1290--1299, 2022{\natexlab{b}}.

\bibitem[Cheng et~al.(2021{\natexlab{b}})Cheng, Nandi, Natarajan, and
  Abd-Almageed]{cheng2021sign}
Cheng, J., Nandi, S., Natarajan, P., and Abd-Almageed, W.
\newblock Sign: Spatial-information incorporated generative network for
  generalized zero-shot semantic segmentation.
\newblock In \emph{ICCV}, pp.\  9556--9566, 2021{\natexlab{b}}.

\bibitem[Contributors(2020)]{mmseg2020}
Contributors, M.
\newblock {MMSegmentation}: Openmmlab semantic segmentation toolbox and
  benchmark.
\newblock \url{https://github.com/open-mmlab/mmsegmentation}, 2020.

\bibitem[Ding et~al.(2022)Ding, Xue, Xia, and Dai]{ding2022decoupling}
Ding, J., Xue, N., Xia, G.-S., and Dai, D.
\newblock Decoupling zero-shot semantic segmentation.
\newblock In \emph{CVPR}, pp.\  11583--11592, 2022.

\bibitem[Dosovitskiy et~al.(2020)Dosovitskiy, Beyer, Kolesnikov, Weissenborn,
  Zhai, Unterthiner, Dehghani, Minderer, Heigold, Gelly,
  et~al.]{dosovitskiy2020image}
Dosovitskiy, A., Beyer, L., Kolesnikov, A., Weissenborn, D., Zhai, X.,
  Unterthiner, T., Dehghani, M., Minderer, M., Heigold, G., Gelly, S., et~al.
\newblock An image is worth 16x16 words: Transformers for image recognition at
  scale.
\newblock \emph{arXiv preprint arXiv:2010.11929}, 2020.

\bibitem[Everingham et~al.(2015)Everingham, Eslami, Van~Gool, Williams, Winn,
  and Zisserman]{everingham2015pascal}
Everingham, M., Eslami, S.~A., Van~Gool, L., Williams, C.~K., Winn, J., and
  Zisserman, A.
\newblock The pascal visual object classes challenge: A retrospective.
\newblock \emph{IJCV}, 111:\penalty0 98--136, 2015.

\bibitem[Ghiasi et~al.(2022)Ghiasi, Gu, Cui, and Lin]{ghiasi2022scaling}
Ghiasi, G., Gu, X., Cui, Y., and Lin, T.-Y.
\newblock Scaling open-vocabulary image segmentation with image-level labels.
\newblock In \emph{ECCV}, pp.\  540--557, 2022.

\bibitem[Gu et~al.(2021)Gu, Lin, Kuo, and Cui]{gu2021open}
Gu, X., Lin, T.-Y., Kuo, W., and Cui, Y.
\newblock Open-vocabulary object detection via vision and language knowledge
  distillation.
\newblock In \emph{ICLR}, 2021.

\bibitem[Gu et~al.(2020)Gu, Zhou, Niu, Zhao, and Zhang]{gu2020context}
Gu, Z., Zhou, S., Niu, L., Zhao, Z., and Zhang, L.
\newblock Context-aware feature generation for zero-shot semantic segmentation.
\newblock In \emph{ACM Multimedia}, pp.\  1921--1929, 2020.

\bibitem[Guo et~al.(2023)Guo, Wang, Gao, Jiang, Tang, Hu, and
  Zhang]{guo2023mvp}
Guo, J., Wang, Q., Gao, Y., Jiang, X., Tang, X., Hu, Y., and Zhang, B.
\newblock Mvp-seg: Multi-view prompt learning for open-vocabulary semantic
  segmentation.
\newblock \emph{arXiv preprint arXiv:2304.06957}, 2023.

\bibitem[Guo et~al.(2022)Guo, Lu, Hou, Liu, Cheng, and Hu]{guo2022segnext}
Guo, M.-H., Lu, C.-Z., Hou, Q., Liu, Z., Cheng, M.-M., and Hu, S.-M.
\newblock Segnext: Rethinking convolutional attention design for semantic
  segmentation.
\newblock In \emph{NeurIPS}, pp.\  1140--1156, 2022.

\bibitem[Han et~al.(2023{\natexlab{a}})Han, Zhong, Li, Han, and
  Ma]{han2023open}
Han, C., Zhong, Y., Li, D., Han, K., and Ma, L.
\newblock Open-vocabulary semantic segmentation with decoupled one-pass
  network.
\newblock In \emph{ICCV}, pp.\  1086--1096, 2023{\natexlab{a}}.

\bibitem[Han et~al.(2023{\natexlab{b}})Han, Liu, Liew, Ding, Liu, Wang, Tang,
  Yang, Feng, Zhao, and Wei]{han2023global}
Han, K., Liu, Y., Liew, J.~H., Ding, H., Liu, J., Wang, Y., Tang, Y., Yang, Y.,
  Feng, J., Zhao, Y., and Wei, Y.
\newblock Global knowledge calibration for fast open-vocabulary segmentation.
\newblock In \emph{ICCV}, pp.\  797--807, 2023{\natexlab{b}}.

\bibitem[He et~al.(2016)He, Zhang, Ren, and Sun]{he2016deep}
He, K., Zhang, X., Ren, S., and Sun, J.
\newblock Deep residual learning for image recognition.
\newblock In \emph{CVPR}, pp.\  770--778, 2016.

\bibitem[Hou et~al.(2020)Hou, Zhang, Cheng, and Feng]{hou2020strip}
Hou, Q., Zhang, L., Cheng, M.-M., and Feng, J.
\newblock Strip pooling: Rethinking spatial pooling for scene parsing.
\newblock In \emph{CVPR}, pp.\  4003--4012, 2020.

\bibitem[Huang et~al.(2021)Huang, Wei, Wang, Liu, Huang, and
  Shi]{huang2021alignseg}
Huang, Z., Wei, Y., Wang, X., Liu, W., Huang, T.~S., and Shi, H.
\newblock Alignseg: Feature-aligned segmentation networks.
\newblock \emph{IEEE TPAMI}, 44\penalty0 (1):\penalty0 550--557, 2021.

\bibitem[Jia et~al.(2021)Jia, Yang, Xia, Chen, Parekh, Pham, Le, Sung, Li, and
  Duerig]{pmlr-v139-jia21b}
Jia, C., Yang, Y., Xia, Y., Chen, Y.-T., Parekh, Z., Pham, H., Le, Q., Sung,
  Y.-H., Li, Z., and Duerig, T.
\newblock Scaling up visual and vision-language representation learning with
  noisy text supervision.
\newblock In \emph{ICML}, pp.\  4904--4916, 2021.

\bibitem[Jiao et~al.(2023)Jiao, Wei, Wang, Zhao, and Shi]{jiao2023learning}
Jiao, S., Wei, Y., Wang, Y., Zhao, Y., and Shi, H.
\newblock Learning mask-aware clip representations for zero-shot segmentation.
\newblock In \emph{NeurIPS}, pp.\  35631--35653, 2023.

\bibitem[Khattak et~al.(2023)Khattak, Wasim, Naseer, Khan, Yang, and
  Khan]{Khattak_2023_ICCV}
Khattak, M.~U., Wasim, S.~T., Naseer, M., Khan, S., Yang, M.-H., and Khan,
  F.~S.
\newblock Self-regulating prompts: Foundational model adaptation without
  forgetting.
\newblock In \emph{ICCV}, pp.\  15190--15200, 2023.

\bibitem[Kim et~al.(2021)Kim, Son, and Kim]{pmlr-v139-kim21k}
Kim, W., Son, B., and Kim, I.
\newblock Vilt: Vision-and-language transformer without convolution or region
  supervision.
\newblock In \emph{ICML}, pp.\  5583--5594, 2021.

\bibitem[Kornblith et~al.(2019)Kornblith, Norouzi, Lee, and
  Hinton]{kornblith2019similarity}
Kornblith, S., Norouzi, M., Lee, H., and Hinton, G.
\newblock Similarity of neural network representations revisited.
\newblock In \emph{ICML}, pp.\  3519--3529, 2019.

\bibitem[Li et~al.(2023)Li, Chen, Qian, and Jia]{li2023tagclip}
Li, J., Chen, P., Qian, S., and Jia, J.
\newblock Tagclip: Improving discrimination ability of open-vocabulary semantic
  segmentation.
\newblock \emph{arXiv preprint arXiv:2304.07547}, 2023.

\bibitem[Li et~al.(2020)Li, Wei, and Yang]{li2020consistent}
Li, P., Wei, Y., and Yang, Y.
\newblock Consistent structural relation learning for zero-shot segmentation.
\newblock In \emph{NeurIPS}, pp.\  10317--10327, 2020.

\bibitem[Lin et~al.(2017)Lin, Goyal, Girshick, He, and
  Doll{\'a}r]{lin2017focal}
Lin, T.-Y., Goyal, P., Girshick, R., He, K., and Doll{\'a}r, P.
\newblock Focal loss for dense object detection.
\newblock In \emph{ICCV}, pp.\  2980--2988, 2017.

\bibitem[Liu et~al.(2022)Liu, Wen, Han, Xu, Xu, and Liang]{liu2022open}
Liu, Q., Wen, Y., Han, J., Xu, C., Xu, H., and Liang, X.
\newblock Open-world semantic segmentation via contrasting and clustering
  vision-language embedding.
\newblock In \emph{ECCV}, pp.\  275--292, 2022.

\bibitem[Loshchilov \& Hutter(2019)Loshchilov and Hutter]{adamw}
Loshchilov, I. and Hutter, F.
\newblock Decoupled weight decay regularization.
\newblock In \emph{ICLR}, 2019.

\bibitem[Luo et~al.(2023)Luo, Bao, Wu, He, and Li]{luo2023segclip}
Luo, H., Bao, J., Wu, Y., He, X., and Li, T.
\newblock Segclip: Patch aggregation with learnable centers for open-vocabulary
  semantic segmentation.
\newblock In \emph{ICML}, pp.\  23033--23044, 2023.

\bibitem[Mikolov et~al.(2013)Mikolov, Sutskever, Chen, Corrado, and
  Dean]{mikolov2013distributed}
Mikolov, T., Sutskever, I., Chen, K., Corrado, G.~S., and Dean, J.
\newblock Distributed representations of words and phrases and their
  compositionality.
\newblock In \emph{NeurIPS}, 2013.

\bibitem[Milletari et~al.(2016)Milletari, Navab, and Ahmadi]{milletari2016v}
Milletari, F., Navab, N., and Ahmadi, S.-A.
\newblock V-net: Fully convolutional neural networks for volumetric medical
  image segmentation.
\newblock In \emph{3DV}, pp.\  565--571, 2016.

\bibitem[Mottaghi et~al.(2014)Mottaghi, Chen, Liu, Cho, Lee, Fidler, Urtasun,
  and Yuille]{mottaghi2014role}
Mottaghi, R., Chen, X., Liu, X., Cho, N.-G., Lee, S.-W., Fidler, S., Urtasun,
  R., and Yuille, A.
\newblock The role of context for object detection and semantic segmentation in
  the wild.
\newblock In \emph{CVPR}, pp.\  891--898, 2014.

\bibitem[Mukhoti et~al.(2023)Mukhoti, Lin, Poursaeed, Wang, Shah, Torr, and
  Lim]{Mukhoti_2023_CVPR}
Mukhoti, J., Lin, T.-Y., Poursaeed, O., Wang, R., Shah, A., Torr, P.~H., and
  Lim, S.-N.
\newblock Open vocabulary semantic segmentation with patch aligned contrastive
  learning.
\newblock In \emph{CVPR}, pp.\  19413--19423, 2023.

\bibitem[Pastore et~al.(2021)Pastore, Cermelli, Xian, Mancini, Akata, and
  Caputo]{Pastore_2021_CVPR}
Pastore, G., Cermelli, F., Xian, Y., Mancini, M., Akata, Z., and Caputo, B.
\newblock A closer look at self-training for zero-label semantic segmentation.
\newblock In \emph{CVPR}, pp.\  2693--2702, 2021.

\bibitem[Qin et~al.(2023)Qin, Wu, Yan, Li, Yuxi, Xiao, Wang, Wang, Wen, Pan,
  and Wang]{Qin_2023_CVPR}
Qin, J., Wu, J., Yan, P., Li, M., Yuxi, R., Xiao, X., Wang, Y., Wang, R., Wen,
  S., Pan, X., and Wang, X.
\newblock Freeseg: Unified, universal and open-vocabulary image segmentation.
\newblock In \emph{CVPR}, pp.\  19446--19455, 2023.

\bibitem[Radford et~al.(2021)Radford, Kim, Hallacy, Ramesh, Goh, Agarwal,
  Sastry, Askell, Mishkin, Clark, Krueger, and Sutskever]{pmlr-v139-radford21a}
Radford, A., Kim, J.~W., Hallacy, C., Ramesh, A., Goh, G., Agarwal, S., Sastry,
  G., Askell, A., Mishkin, P., Clark, J., Krueger, G., and Sutskever, I.
\newblock Learning transferable visual models from natural language
  supervision.
\newblock In \emph{ICML}, pp.\  8748--8763, 2021.

\bibitem[Rao et~al.(2022)Rao, Zhao, Chen, Tang, Zhu, Huang, Zhou, and
  Lu]{Rao_2022_CVPR}
Rao, Y., Zhao, W., Chen, G., Tang, Y., Zhu, Z., Huang, G., Zhou, J., and Lu, J.
\newblock Denseclip: Language-guided dense prediction with context-aware
  prompting.
\newblock In \emph{CVPR}, pp.\  18082--18091, 2022.

\bibitem[Shin et~al.(2022)Shin, Xie, and Albanie]{shin2022reco}
Shin, G., Xie, W., and Albanie, S.
\newblock Reco: Retrieve and co-segment for zero-shot transfer.
\newblock In \emph{NeurIPS}, pp.\  33754--33767, 2022.

\bibitem[Sun et~al.(2023)Sun, Li, Torr, Gu, and Li]{sun2023clip}
Sun, S., Li, R., Torr, P., Gu, X., and Li, S.
\newblock Clip as rnn: Segment countless visual concepts without training
  endeavor.
\newblock \emph{arXiv preprint arXiv:2312.07661}, 2023.

\bibitem[Wang et~al.(2022)Wang, Lu, Li, Tao, Guo, Gong, and
  Liu]{Wang_2022_CVPR}
Wang, Z., Lu, Y., Li, Q., Tao, X., Guo, Y., Gong, M., and Liu, T.
\newblock Cris: Clip-driven referring image segmentation.
\newblock In \emph{CVPR}, pp.\  11686--11695, 2022.

\bibitem[Xian et~al.(2019)Xian, Choudhury, He, Schiele, and
  Akata]{xian2019semantic}
Xian, Y., Choudhury, S., He, Y., Schiele, B., and Akata, Z.
\newblock Semantic projection network for zero-and few-label semantic
  segmentation.
\newblock In \emph{CVPR}, pp.\  8256--8265, 2019.

\bibitem[Xie et~al.(2021)Xie, Wang, Yu, Anandkumar, Alvarez, and
  Luo]{xie2021segformer}
Xie, E., Wang, W., Yu, Z., Anandkumar, A., Alvarez, J.~M., and Luo, P.
\newblock Segformer: Simple and efficient design for semantic segmentation with
  transformers.
\newblock In \emph{NeurIPS}, pp.\  12077--12090, 2021.

\bibitem[Xu et~al.(2022{\natexlab{a}})Xu, De~Mello, Liu, Byeon, Breuel, Kautz,
  and Wang]{Xu_2022_CVPR}
Xu, J., De~Mello, S., Liu, S., Byeon, W., Breuel, T., Kautz, J., and Wang, X.
\newblock Groupvit: Semantic segmentation emerges from text supervision.
\newblock In \emph{CVPR}, pp.\  18134--18144, 2022{\natexlab{a}}.

\bibitem[Xu et~al.(2023)Xu, Hou, Zhang, Feng, Wang, Qiao, and
  Xie]{xu2023learning}
Xu, J., Hou, J., Zhang, Y., Feng, R., Wang, Y., Qiao, Y., and Xie, W.
\newblock Learning open-vocabulary semantic segmentation models from natural
  language supervision.
\newblock In \emph{CVPR}, pp.\  2935--2944, 2023.

\bibitem[Xu et~al.(2022{\natexlab{b}})Xu, Zhang, Wei, Lin, Cao, Hu, and
  Bai]{xu2021simple}
Xu, M., Zhang, Z., Wei, F., Lin, Y., Cao, Y., Hu, H., and Bai, X.
\newblock A simple baseline for zero-shot semantic segmentation with
  pre-trained vision-language model.
\newblock In \emph{ECCV}, pp.\  736--753, 2022{\natexlab{b}}.

\bibitem[Xu et~al.(2024)Xu, Xu, Wang, Xu, Guo, Zhang, and
  Zhang]{xu2023spectral}
Xu, W., Xu, R., Wang, C., Xu, S., Guo, L., Zhang, M., and Zhang, X.
\newblock Spectral prompt tuning: Unveiling unseen classes for zero-shot
  semantic segmentation.
\newblock In \emph{AAAI}, pp.\  6369--6377, 2024.

\bibitem[Zeng et~al.(2022)Zeng, Zhang, and Li]{zeng2022multi}
Zeng, Y., Zhang, X., and Li, H.
\newblock Multi-grained vision language pre-training: Aligning texts with
  visual concepts.
\newblock In \emph{ICML}, pp.\  25994--26009, 2022.

\bibitem[Zhang et~al.(2022)Zhang, Tian, Tang, Chu, Wei, Shen,
  et~al.]{zhang2022segvit}
Zhang, B., Tian, Z., Tang, Q., Chu, X., Wei, X., Shen, C., et~al.
\newblock Segvit: Semantic segmentation with plain vision transformers.
\newblock In \emph{NeurIPS}, pp.\  4971--4982, 2022.

\bibitem[Zhao et~al.(2017)Zhao, Shi, Qi, Wang, and Jia]{Zhao_2017_CVPR}
Zhao, H., Shi, J., Qi, X., Wang, X., and Jia, J.
\newblock Pyramid scene parsing network.
\newblock In \emph{CVPR}, 2017.

\bibitem[Zheng et~al.(2021)Zheng, Lu, Zhao, Zhu, Luo, Wang, Fu, Feng, Xiang,
  Torr, et~al.]{zheng2021rethinking}
Zheng, S., Lu, J., Zhao, H., Zhu, X., Luo, Z., Wang, Y., Fu, Y., Feng, J.,
  Xiang, T., Torr, P.~H., et~al.
\newblock Rethinking semantic segmentation from a sequence-to-sequence
  perspective with transformers.
\newblock In \emph{CVPR}, pp.\  6881--6890, 2021.

\bibitem[Zhou et~al.(2022{\natexlab{a}})Zhou, Loy, and Dai]{zhou2022extract}
Zhou, C., Loy, C.~C., and Dai, B.
\newblock Extract free dense labels from clip.
\newblock In \emph{ECCV}, pp.\  696--712, 2022{\natexlab{a}}.

\bibitem[Zhou et~al.(2022{\natexlab{b}})Zhou, Yang, Loy, and
  Liu]{zhou2022learning}
Zhou, K., Yang, J., Loy, C.~C., and Liu, Z.
\newblock Learning to prompt for vision-language models.
\newblock \emph{IJCV}, 130\penalty0 (9):\penalty0 2337--2348,
  2022{\natexlab{b}}.

\bibitem[Zhou et~al.(2023)Zhou, Lei, Zhang, Liu, and Liu]{Zhou_2023_CVPR}
Zhou, Z., Lei, Y., Zhang, B., Liu, L., and Liu, Y.
\newblock Zegclip: Towards adapting clip for zero-shot semantic segmentation.
\newblock In \emph{CVPR}, pp.\  11175--11185, 2023.

\bibitem[Zhu \& Chen(2023)Zhu and Chen]{zhu2023survey}
Zhu, C. and Chen, L.
\newblock A survey on open-vocabulary detection and segmentation: Past,
  present, and future.
\newblock \emph{arXiv preprint arXiv:2307.09220}, 2023.

\end{thebibliography}
\bibliographystyle{icml2024}

\newpage
\clearpage
\appendix
\section{Implementation Details}
\myPara{Text templates for prompts.}
Following the training details of CLIP, we utilize text templates to generate embeddings for class descriptions using the CLIP text encoder.
Specifically, we employ a single template, "A photo of a \{ \}," for PASCAL VOC 2012 (VOC). 
For large-scale datasets such as COCO-Stuff 164K (COCO) and PASCAL Context (Context), we employ 15 augmented templates for improved representation.
The details of the 15 augmented templates are: 
`A photo of a \{ \}.', 
`A photo of a small \{ \}.',
`A photo of a medium \{ \}.',
`A photo of a large \{ \}.',
`This is a photo of a \{ \}.',
`This is a photo of a small \{ \}.',
`This is a photo of a medium \{ \}.',
`This is a photo of a large \{ \}.',
`A \{ \} in the scene.',
`A photo of a \{ \} in the scene.',
`There is a \{ \} in the scene.',
`There is the \{ \} in the scene.',
`This is a \{ \} in the scene.',
`This is the \{ \} in the scene.' and
`This is one \{ \} in the scene.'.

\section{Datasets and Evaluation Metrics}
\label{A-sec: dataset}
\myPara{Datasets.}
\textbf{COCO-Stuff} is an extensive semantic segmentation dataset comprising 171 categories, encompassing 80 things classes and 91 stuff classes. 
It contains 117k training images and 5k validation images and it is divided into 156 seen classes and 15 unseen classes.
In comparison, \textbf{PASCAL VOC} consists of 11,185 training images and 1,449 validation images across 20 classes. 
We exclude the `background' category, utilizing 15 classes as the seen part and 5 classes as the unseen part.
Additionally, \textbf{PASCAL Context} provides supplementary annotations for PASCAL VOC 2010, consisting of 4,998 training images and 5,005 validation images.
For testing, we classify the dataset into 49 seen classes (excluding `background') and use the remaining 10 classes as unseen.

\myPara{Data split.}
We adopt the identical unseen class setup introduced in the previous method~\cite{Zhou_2023_CVPR} to ensure a fair comparison.
The specific name of unseen classes for COCO-Stuff~(COCO), PASCAL VOC 2012~(VOC), and PASCAL Context~(Context) datasets in~\tabref{tab: dataset-split}.

\begin{table}[!h]
    \centering
    \caption{Details of unseen class names.
    %
    }
    \label{tab: dataset-split}
    \begin{center}
    \begin{small}
    \setlength{\tabcolsep}{1.7mm}{
    \begin{tabular}{lc}
    \toprule
    Dataset & Name\\    
    \midrule
    \multirow{3}{*}{COCO}  & $cow, giraffe, suitcase, frisbee, skateboard,$ \\
      & $carrot, scissors, cardboard, clouds, grass,$ \\
     & $playingfield, river, road, tree, wall concrete$ \\
    \midrule
    VOC & $pottedplant, sheep, sofa, train, tvmonitor$\\
    \midrule
    \multirow{2}{*}{Context} & $cow, motorbike, sofa, cat, boat, fence$\\
    & $bird, tv monitor, keyboard, aeroplane$\\
   \bottomrule
    \end{tabular}}
    \end{small}
    \end{center}
    \vskip -0.1in
\end{table}

\myPara{Evaluation metrics.}
Following the previous work~\cite{Zhou_2023_CVPR,xu2023spectral}, we report the mean IoU (mIoU) for seen and unseen classes denoted as mIoU$^S$ and mIoU$^U$ respectively.
In addition, we compute the harmonic mean IoU (hIoU) among seen and unseen classes, which is calculated as
\begin{equation}
\mathrm{hIoU}=\frac{2(\mathrm{mIoU}^S+\mathrm{mIoU}^U)}{\mathrm{mIoU}^S+\mathrm{mIoU}^U}.
   \end{equation}
The mIoU metric is also adopted for the evaluation of cross-dataset.

\begin{table}[t]
    \centering
    \vskip -0.1in
    \caption{Efficiency comparison with different methods. 
    The input image is set to $512 \times 512.$
    All models are evaluated on a single 3090 GPU. 
    Params. represents the number of parameters of the model.
    Note that our \ourMthd~requires less learnable tokens.
    }
    \label{tab: param-flops}
    \begin{center}
    \begin{small}
    \setlength{\tabcolsep}{0.2mm}{
    \begin{tabular}{lccccc}
    \toprule
    \multirow{2}{*}{Methods}& \multicolumn{2}{c}{Param. (M)}& & \\  
    \cmidrule(lr){2-3} 
     & Total$\downarrow$ &Trainable$\downarrow$ & GFLOPs$\downarrow$& FPS$\uparrow$ \\    
    \midrule
    ZegFormer~\cite{ding2022decoupling} & 210.0& 60.3 & 1875.1& 7.4\\
    DeOP~\cite{han2023open} & 218.4& - & 670.0& 5.8\\
    ZegCLIP~\cite{Zhou_2023_CVPR}&164.9 & 14.6 & 123.9& 20.9 \\
    \ourMthd~(Ours) &193.1& 40.5& 123.8& 18.4\\
    \bottomrule
    \end{tabular}}
    \end{small}
    \end{center}
    \vskip -0.1in
\end{table}

\section{Experiment}
\label{sec: a-exper}
\myPara{Efficiency analysis.}
Besides performance comparison, we also compare our approach with the typical two-encoder methods (\eg ZegFormer~\cite{ding2022decoupling} and DeOP~\cite{han2023open} ) and one-encoder methods (\eg ZegCLIP~\cite{Zhou_2023_CVPR}).
We test these methods under the same environment to ensure a fair comparison, including hardware and image resolution.
\tabref{tab: param-flops} shows that our approach introduces the trainable parameters compared to ZegCLIP, notably less than ZegFormer's.
Furthermore, the total parameter of the model increase is marginal.
Moreover, our \ourMthd~ does not increase the computational effort since \ourMthd~requires fewer learnable tokens than ZegCLIP (refer to subsequent experiments for details).
Because of this, our \ourMthd~can still maintain a high FPS speed during inference.



\begin{table}[!t]
    \centering
    \vskip -0.1in
    \caption{Generalization ability to other datasets.
    The models are trained on the COCO-Stuff dataset.
    }
    \label{tab: cross-dataset}
    \begin{center}
    \begin{footnotesize}
    \setlength{\tabcolsep}{0.1mm}{
    \begin{tabular}{lccc}
    \toprule
    Methods & Backbone & VOC$\uparrow$& Context$\uparrow$ \\    
    \midrule
    \multicolumn{1}{l}{\emph{\textbf{Fully Supervised}}} \\
    ZegFormer~\cite{ding2022decoupling}&R101\&CLIP-B& 89.5& 45.5 \\
    Zsseg~\cite{xu2021simple}&R101\&CLIP-B& -& 47.7 \\
    DeOP~\cite{han2023open}&R101\&CLIP-B&91.7&48.8\\
    GroupViT~\cite{Xu_2022_CVPR}& ViT-S& 52.3& 22.4 \\
    LSeg+~\cite{ghiasi2022scaling}& R101& 59.0& 36.0 \\
    ViL-Seg~\cite{liu2022open}& ViT-B&33.6&15.9 \\
    SegCLIP~\cite{luo2023segclip}&CLIP-B& 52.6& 24.7 \\
    OpenSeg~\cite{ghiasi2022scaling}&R101& 60.0& 36.9 \\
    OVSegmentor~\cite{xu2023learning}& ViT-B& 53.8& 20.4 \\
    PACL~\cite{Mukhoti_2023_CVPR} & ViT-B& 72.3& 50.1 \\
    TCL~\cite{Cha_2023_CVPR}&CLIP-B &83.2&33.9\\
    GKC~\cite{han2023global}& R101&83.2&45.2 \\
    \midrule
    \multicolumn{1}{l}{\emph{\textbf{Inductive}}} \\
    ZegCLIP~\cite{Zhou_2023_CVPR}& CLIP-B & 93.6& 47.5\\
    \ourMthd~(Ours)& CLIP-B &93.8 &48.4\\
    \midrule
    \multicolumn{1}{l}{\emph{\textbf{Transductive}}} \\
    ZegCLIP+ST~\cite{Zhou_2023_CVPR}& CLIP-B & \textbf{94.1}& 52.7\\
    \ourMthd+ST~(Ours) & CLIP-B& 94.0& \textbf{52.8}\\

    \bottomrule
    \end{tabular}}
    \end{footnotesize}
    \end{center}
    \vskip -0.1in
\end{table}

\begin{table}[!t]
    \centering
    \small
    \setlength{\abovecaptionskip}{5pt}
    \setlength{\tabcolsep}{0.7mm}
    \caption{Effect of applying the different numbers of learnable tokens.
    We report the results of ZegCLIP and our \ourMthd.
    }
    \label{tab: learnable-tokens}
    \begin{tabular}{lcccccccc} \toprule
    & \multicolumn{3}{c}{ZegCLIP}&\multicolumn{3}{c}{\ourMthd~(Ours)} \\  
    \cmidrule(lr){2-4}  \cmidrule(lr){5-7}
    Number & mIoU$^S$$\uparrow$& mIoU$^U$$\uparrow$& hIoU$\uparrow$& mIoU$^S$$\uparrow$& mIoU$^U$$\uparrow$& hIoU$\uparrow$ \\ \midrule
    35  &39.4 &40.9 & 40.1& 40.6& 42.3& 41.4\\
    50  &39.2 &39.7 & 39.5& \textbf{41.1} & \textbf{43.4} & \textbf{42.2} \\
    100 &\textbf{40.2} &\textbf{41.4} & \textbf{40.8}& 40.8& 42.0& 41.4 \\
    \bottomrule
    \end{tabular}
    \vskip -0.1in
\end{table}

\myPara{Generalization ability to other datasets.}
To further explore the generalization ability of our proposed \ourMthd, we perform additional experiments detailed in~\tabref{tab: cross-dataset}.
In this setting, we utilize the pre-trained model from the source dataset (\ie COCO) through supervised learning on seen classes, assessing segmentation performance on both seen and unseen classes in the target datasets (\ie VOC and Context).
We also compare the previous state-of-the-art zero-shot segmentation methods 
including the two-encoder approaches (\eg ZegFormer~\cite{ding2022decoupling}, Zsseg~\cite{xu2021simple} and DeOP~\cite{han2023open}) 
and one-encoder approaches (\eg GKC~\cite{han2023global} and TCL~\cite{Cha_2023_CVPR}).
Our approach exhibits superior cross-domain generalization compared to the previous methods, particularly outperforming the recent ZegCLIP, which is also built on the CLIP model.

\myPara{Effect of the number of learnable tokens.}
The learnable tokens are integrated into the CLIP frozen visual encoder to facilitate deep prompt tuning.
We present results for different numbers of learnable tokens and compare them against the baseline ZegCLIP.
As shown in~\tabref{tab: learnable-tokens}, our proposed \ourMthd~with 35 learnable tokens outperforms ZegCLIP with 100 tokens, achieving the best performance with 50 learnable tokens.
This demonstrates the superior effectiveness and efficiency of our approach in zero-shot segmentation by leveraging multi-level visual features in a cascade manner.


\begin{table}[t]
    \centering
    \caption{Ablation of cascaded decoder.
    \cmark indicates the fusion of \{ \}-layer features in the Transformer block, aligning them with an individual decoder.
    }
    \label{tab: effect-seg-encoder}
    \begin{center}
    \begin{small}
    \setlength{\tabcolsep}{2.0mm}{
    \begin{tabular}{cccc ccc}
    \toprule
    \{6-8\} & \{9-11\}& \{12\}& mIoU$^S$$\uparrow$& mIoU$^U$$\uparrow$& hIoU$\uparrow$ \\    
    \midrule
    &\cmark & \cmark & 40.2 & 40.2 & 40.2 \\
    \cmark & & \cmark & 40.5 & 42.2 & 41.3 \\
      \cmark &\cmark & & 40.3 & 41.8 & 41.1\\
      \midrule
      \cmark &\cmark & \cmark & \textbf{41.1} & \textbf{43.4} & \textbf{42.2} \\

    \bottomrule
    \end{tabular}}
    \end{small}
    \end{center}
    \vskip -0.1in
\end{table}
\begin{table}[!t]
    \centering
    \caption{Ablation of variance $\sigma$ of the Gaussian function in NGA module. We present the initial weights of the corresponding features at different variance values.
    }
    \label{tab: gaussian-para}
    \begin{center}
    \begin{small}
    \setlength{\tabcolsep}{2.2mm}{
    \begin{tabular}{ccccc}
    \toprule
    $\sigma$ & values of weights & mIoU$^S$$\uparrow$& mIoU$^U$$\uparrow$& hIoU$\uparrow$ \\
    \midrule
     0.8& \{0.04, 0.46, 1.00\}&  92.3& 82.6 &87.3\\
     1.0& \{0.14, 0.61, 1.00\}& \textbf{92.7}&\textbf{83.1}&\textbf{87.7}\\
     1.2& \{0.25, 0.71, 1.00\} & 92.4& 79.1 & 85.1\\
    \bottomrule
    \end{tabular}}
    \end{small}
    \end{center}
    \vskip -0.2in
\end{table}
\begin{table}[!t]
    \centering
    \caption{Variance weights of the Gaussian function in NGA module before and after training.
    }
    \label{tab: gaussian-train}
    \begin{center}
    \begin{small}
    \setlength{\tabcolsep}{.1mm}{
    \begin{tabular}{lcccc}
    \toprule
    Block splitting manner& Initial weights & Weight after training\\
    \midrule
    \{6-8\}& \{0.14, 0.61, 1.00\}& \{0.13, 0.59, 1.02\} \\
    \{9-11\}& \{0.14, 0.61, 1.00\}& \{0.13, 0.59, 1.01\}\\
    \bottomrule
    \end{tabular}}
    \end{small}
    \end{center}
    \vskip -0.2in
\end{table}
\begin{table}[!t]
    \centering
    \caption{Ablation of weights of dice loss ($\alpha$) and focal loss ($\beta$).
    }
    \label{tab: loss-weights}
    \begin{center}
    \begin{small}
    \setlength{\tabcolsep}{4.2mm}{
    \begin{tabular}{ccccc}
    \toprule
    $\alpha$ & $\beta$ & mIoU$^S$$\uparrow$& mIoU$^U$$\uparrow$& hIoU$\uparrow$ \\
    \midrule
     1& 10&87.7& 72.4&79.3\\
     1& 50&91.3&81.1&85.9\\
     1& 100&\textbf{92.7}&\textbf{83.1}&\textbf{87.7}\\
     1& 150&91.3&82.5&86.7\\
    \bottomrule
    \end{tabular}}
    \end{small}
    \end{center}
    \vskip -0.2in
\end{table}

\myPara{Ablation on the cascaded encoders.}
Our \ourMthd~can flexibly integrate the features from different encoder blocks.
As shown in~\tabref{tab: effect-seg-encoder}, both deep (the 1st result) and intermediate features (the 2nd result) contribute to improving the performance compared to using only the last features.
This again demonstrates that intermediate layer features can complement last layer features, thus improving segmentation results.
Notably, competitive performance is achieved even without employing last-layer features (the 3rd result).
Moreover, simultaneously fusing these features further enhances the overall model performance,
indicating that our approach has the ability to effectively exploit distinctive features without compromising transferability.

\myPara{Ablation on the variance parameter in NGA.}
Our NGA module assigns distinct Gaussian weights to blocks for feature fusion based on their neighborhood relative distances.
The variance $\sigma$ of Gaussian weights is a key parameter:
higher values lead to consistent weighting across feature blocks, while lower values degenerate into reliance on a single layer of features.
\tabref{tab: gaussian-para} shows that increasing the value of $\sigma$ causes the mIoU in the unseen class to drop significantly despite similar performance in seen classes.
We also present the variance weights of the NGA module in different split encoder stages before and after training, as shown in~\tabref{tab: gaussian-train}.

\begin{figure*}[!t]
\centering
    \vspace{5pt}
    \begin{overpic}[width=0.7\textwidth]{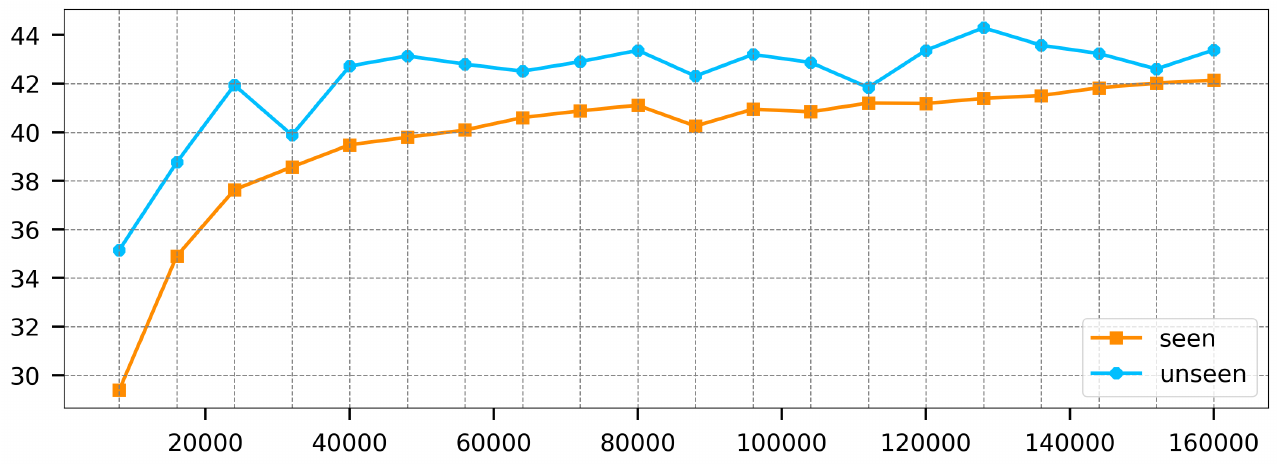}
    \put(-3,20){\rotatebox[]{90}{{mIoU(\%)}}}
    \put(40,-2.5){{Training iterations}}
    \end{overpic}
    \vspace{7pt}
\caption{Intersection over mIoU (\%) on seen and unseen classes rises with the number of training iterations.}
\label{iterations_increase}
    \vskip -0.1in
\end{figure*}

\myPara{Ablation on the loss weights.}
We conduct ablation experiments to examine the impact of dice loss weights ($\alpha$) and focal loss weights ($\beta$) on performance using the Pascal-VOC dataset. 
The results in~\tabref{tab: loss-weights} indicate that emphasizing focal loss weights 100 times more than dice loss weights is more effective.

\myPara{Training iterations.}
We employ 80000 default training iterations on COCO for \ourMthd{} to ensure a fair comparison with previous methods.
Additionally, We investigate the impact of training iterations for our \ourMthd.
\figref{iterations_increase} shows the increment in iterations correlates with a steady enhancement of mIoU for seen classes, accompanied by a notable improvement in performance for unseen classes.
This indicates that our approach maintains strong zero-shot capability on unseen classes, transitioning CLIP from image-level recognition to pixel-level segmentation.

\section{Visualizations}
\label{A-sec: visualization}



\begin{figure*}[t]
\centering
    \begin{overpic}[width=0.9\textwidth]{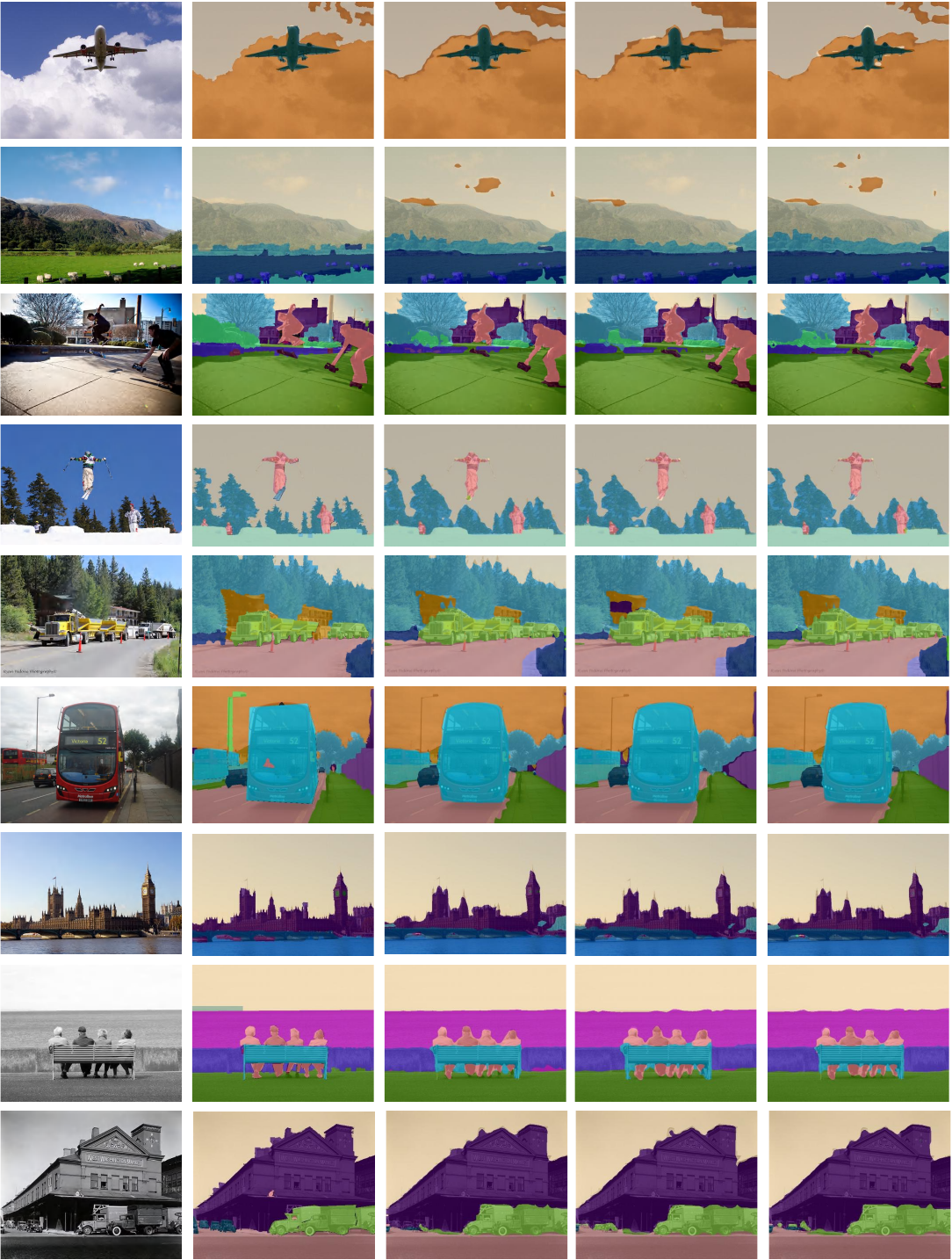}
    \put(3,-2){\footnotesize{Origin Image}}
    \put(18,-2){\footnotesize{Ground Truth}}
    \put(34,-2){\footnotesize{ZegCLIP}}
    \put(50,-2){\footnotesize{SPT-SEG}}
    \put(62,-2){\footnotesize{\ourMthd~(Ours)}}
    \end{overpic}
    \vspace{5pt}
\caption{Comparing qualitative zero-shot segmentation results among various semantic segmentation methods with the transductive setting on the COCO-Stuff dataset.}
\label{A-visualisation_transductive}
\end{figure*}
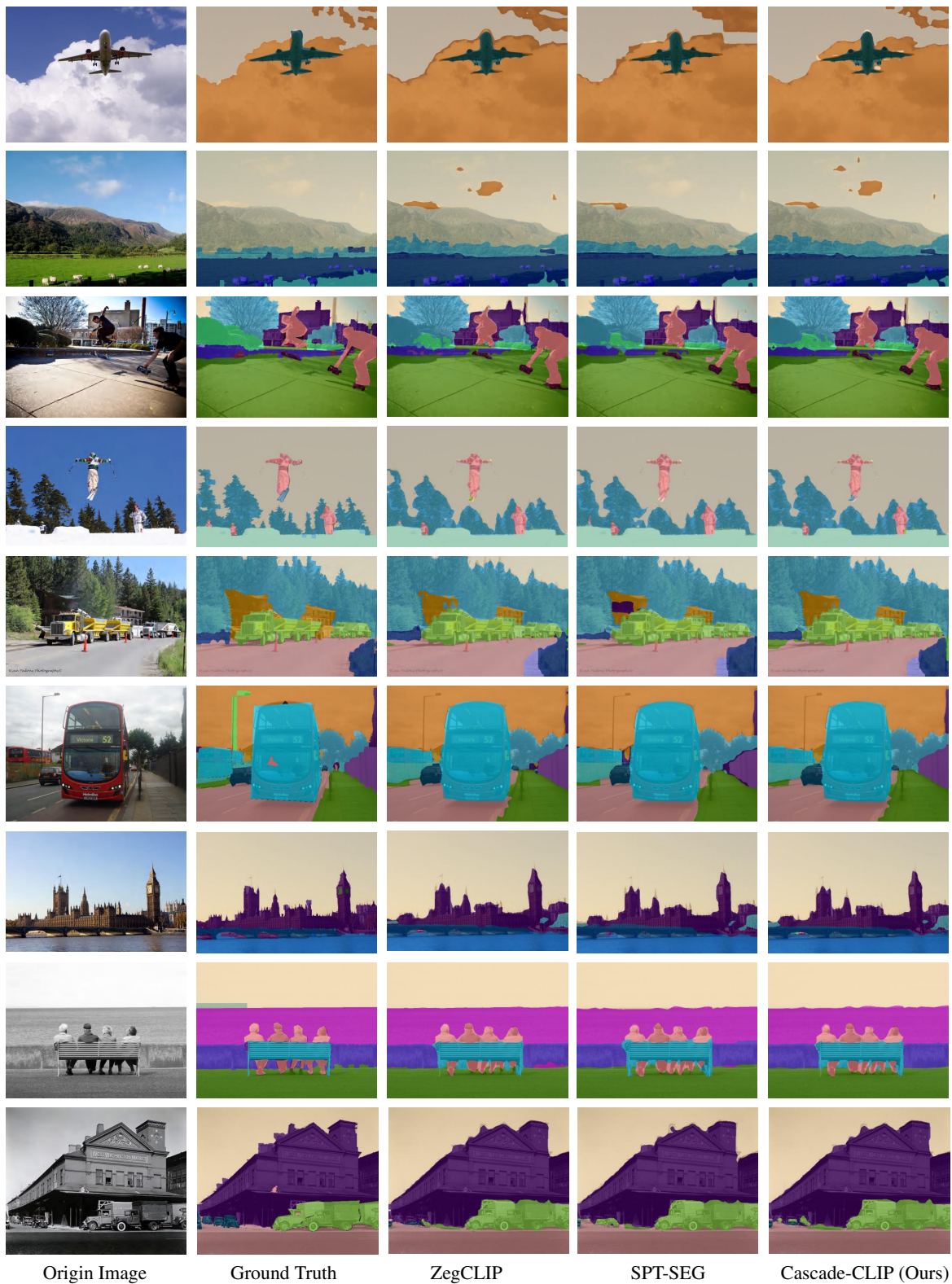

We further evaluate the segmentation performance of our method in transductive settings, comparing it with existing segmentation methods.
The qualitative zero-shot segmentation results shown in~\figref{A-visualisation_transductive} demonstrate that our \ourMthd~model yields very clear segmentation masks in most cases.
For example, our model more accurately segments the shape of the clouds, although their shape is irregular in both the first and second examples.
In the third example, our segmentation mask clearly displays the shape of the skateboard, although it occupies only a very small area in the image.
These observations confirm the remarkable effectiveness of our \ourMthd~approach.

\end{document}